\pdfoutput=1

\documentclass[11pt]{article}

\usepackage[final]{acl}

\usepackage{times}
\usepackage{latexsym}

\usepackage[T1]{fontenc}

\usepackage[utf8]{inputenc}

\usepackage{microtype}
\usepackage{amsmath}

\usepackage{inconsolata}

\usepackage{graphicx}
\usepackage{booktabs}
\usepackage{multirow}
\usepackage{xcolor,colortbl}
\definecolor{Gray}{gray}{0.85}
\usepackage[inkscapelatex=false]{svg}
\newcommand{\ignore}[1]{}

\title{\texttt{M$^3$FinMeeting}: A Multilingual, Multi-Sector, and Multi-Task Financial Meeting Understanding Evaluation Dataset}

\author{Jie Zhu$^{1,2}$, Junhui Li$^1$\thanks{Corresponding Author}, Yalong Wen$^{2}$, Xiandong Li${^3}$, Lifan Guo$^2$, Feng Chen$^2$\\
$^1$School of Computer Science and Technology, Soochow University \\
$^2$Qwen DianJin Team, Alibaba Cloud Computing\\
$^3$Nanjing University \\
{zhujie951121@gmail.com},
{lijunhui@suda.edu.cn} \\
{\{lifan.lg, betterman.chenf\}@alibaba-inc.com}
}

\begin{document}
\maketitle
\begin{abstract}
Recent breakthroughs in large language models (LLMs) have led to the development of new benchmarks for evaluating their performance in the financial domain. However, current financial benchmarks often rely on news articles, earnings reports, or announcements, making it challenging to capture the real-world dynamics of financial meetings. To address this gap, we propose a novel benchmark called \texttt{M$^3$FinMeeting}, which is a multilingual, multi-sector, and multi-task dataset designed for financial meeting understanding. First, \texttt{M$^3$FinMeeting} supports English, Chinese, and Japanese, enhancing comprehension of financial discussions in diverse linguistic contexts. Second, it encompasses various industry sectors defined by the Global Industry Classification Standard (GICS), ensuring that the benchmark spans a broad range of financial activities. Finally, \texttt{M$^3$FinMeeting} includes three tasks: summarization, question-answer (QA) pair extraction, and question answering, facilitating a more realistic and comprehensive evaluation of understanding. Experimental results with seven popular LLMs reveal that even the most advanced long-context models have significant room for improvement, demonstrating the effectiveness of \texttt{M$^3$FinMeeting} as a benchmark for assessing LLMs' financial meeting comprehension skills.\footnote{We make our dataset and project available on \url{https://github.com/aliyun/qwen-dianjin}.}
\end{abstract}

\section{Introduction}

Financial meetings, whether in person or virtual, serve as critical venues for decision-making, negotiation, and strategy formulation among various participants. Although large language models (LLMs) have demonstrated impressive performance across multiple natural language processing (NLP) tasks~\cite{openai-2023-gpt4}, it remains uncertain how they can effectively understand and process lengthy speech texts to help financial professionals expedite their work. Key capabilities, such as summarizing crucial points, responding to inquiries, and extracting question-answer pairs, are particularly beneficial for enhancing productivity and facilitating informed discussions in this context. 

In the financial domain, there are various benchmarks available in different languages, such as FinQA~\cite{chen-etal-2021-finqa} and ConvFinQA~\cite{chen-etal-2022-convfinqa} in English, as well as CFLUE~\cite{zhu-etal-2024-cflue} and the CCKS series shared tasks in Chinese~\cite{tianchi_2019_ccks_entity,tianchi_2020_ccks_entity,tianchi_2021_ccks_causal,tianchi_2022_ccks}\ignore{, as well as ArBanking77\footnote{\url{https://sina.birzeit.edu/arbanking77/arafinnlp/}} in Arabic}. However, these datasets are primarily sourced from financial news and earnings reports, lacking content from real-world financial meetings. Additionally, they are monolingual, limited to English or Chinese. To address this gap, we introduce a new dataset called \texttt{M$^3$FinMeeting}, designed for multilingual and multi-sector evaluation of financial meeting understanding, featuring multiple tasks. First, \texttt{M$^3$FinMeeting} supports multiple languages, including English, Chinese and Japanese, enhancing the understanding of financial discussions across different linguistic contexts. Second, it encompasses all 11 industry sectors defined by Global Industry Classification Standard (GICS). Finally, \texttt{M$^3$FinMeeting} includes multiple classical NLP tasks such as summarization, question answering, and question-answer pair extraction. Importantly, since financial meetings typically last one to two hours, \texttt{M$^3$FinMeeting} provides long-context data, which is essential for assessing the ability of LLMs to handle complex tasks. Additionally, the tasks within \texttt{M$^3$FinMeeting} require LLMs to generate long responses, allowing for a thorough assessment of their capabilities in producing coherent and relevant outputs in challenging scenarios.

Based on \texttt{M$^3$FinMeeting}, we assess the effectiveness of seven representative LLMs including two OpenAI GPTs and five open-sourced LLMs. The experimental results indicate that Qwen2.5-72B-Instruct~\cite{yang-etal-2024-qwen2.5} significantly outperforms other large language models (LLMs), achieving overall scores above 70 when evaluated by GPT-4~\cite{openai-2023-gpt4}. However, this also suggests that even the most advanced LLMs currently available struggle with the tasks in \texttt{M$^3$FinMeeting}, revealing substantial rooms for performance improvement.

Our main contributions can be summarized as follows:
\begin{itemize}
\item \texttt{M$^3$FinMeeting} introduces a novel evaluation benchmark specifically designed for financial meetings, addressing the lack of real-world financial meeting data in existing benchmarks.
\item \texttt{M$^3$FinMeeting} supports multilingual evaluation in English, Chinese, and Japanese, spans the 11 industry sectors defined by GICS, and includes three key NLP tasks: summarization, question-answer pair extraction, and question answering. All documents are carefully annotated by financial analysts to ensure high-quality and accurate evaluation\ignore{, making this the first benchmark designed for summarization and question answering based on real-world meetings}.
\item Through extensive experiments and detailed analyses, we thoroughly assess the performance of state-of-the-art long-context LLMs, offering valuable insights into their limitations and potential improvements for better long-context modeling in financial scenarios.
\end{itemize}

\section{Related Work}
\label{sec:related_work}

\begin{table*}[!t]
\centering
\small
\resizebox{\textwidth}{!}{
\begin{tabular}{l|l|l|l}
\toprule
\bf Language & \bf Dataset & \bf Sources & \bf Tasks\\
\midrule
\multirow{8}{*}{English} & FinQA~\cite{chen-etal-2021-finqa} & Earnings reports & Question answering \& reasoning\\
\cmidrule{2-4}
& ConvFinQA~\cite{chen-etal-2022-convfinqa} & Reports & Question answering \& reasoning\\
\cmidrule{2-4}
& FLUE~\cite{shah-etal-2022-flue} & Multiple & 5 tasks\\
\cmidrule{2-4}
& ECTSum~\cite{mukherjee-etal-2022-ectsum} & Call transcripts & Summarization\\
\cmidrule{2-4}
& FLARE~\cite{xie-etal-2023-pixiu} & Multiple & 8 tasks\\
\cmidrule{2-4}
& FinTextQA~\cite{chen-etal-2024-fintextqa} & Long reports & Question answering\\
\cmidrule{2-4}
& BizBench~\cite{krumdick-etal-2024-bizbench} & Multiple & 8 quantitative reasoning tasks \\
\cmidrule{2-4}
& FinanceBench~\cite{islam-etal-2023-financebench}  & \begin{tabular}[c]{@{}l@{}}Public filings, \\ Earnings reports\end{tabular} & Question answering \\
\midrule
\multirow{7}{*}{Chinese} 
& CCKS~\citet{tianchi_2019_ccks_entity,tianchi_2020_ccks_entity,tianchi_2021_ccks_causal,tianchi_2022_ccks} & Multiple & Multiple event-related tasks \\ 
\cmidrule{2-4}
& DCFEE~\cite{yang-etal-2018-dcfee} & Announcements & Document-level event extraction\\
\cmidrule{2-4}
& Doc2EDAG~\cite{zheng-etal-2019-doc2edag} & Announcements 
 & Document-level event extraction\\
\cmidrule{2-4}
& DuEE-Fin~\cite{han-etal-2022-dueefin} & Multiple \ignore{\begin{tabular}[c]{@{}l@{}}Announcements, \\judgements,\\ news articles\end{tabular}} & Document-level event extraction\\
\cmidrule{2-4}
& CFinDEE~\cite{zhang-etal-2024-cfindee} & News articles & Document-level event extraction\\
\cmidrule{2-4}
& BBT-CFLEB~\cite{lu-etal-2023-bbtfin} & Multiple & 6 tasks\\
\cmidrule{2-4}
& CFLUE~\cite{zhu-etal-2024-cflue} & Multiple & \begin{tabular}[c]{@{}l@{}} Multiple-choice Question Answering\\ \& Reasoning and 5 tasks\end{tabular} \\
\midrule
\begin{tabular}[c]{@{}l@{}}English, \\ Chinese,\\ Japanese\end{tabular} & \texttt{M$^3$FinMeeting} (Ours) & Meetings & \begin{tabular}[c]{@{}l@{}} Summarization, \\ Question answering, \\ QA pair extraction \end{tabular} \\
\bottomrule
\end{tabular}
}
\caption{Summary of recent financial benchmarks.}
\label{tab:benchmarks}
\end{table*}

\subsection{Financial Evaluation Benchmarks}

Financial NLP has become a key application area for LLMs, attracting increasing attention for its benchmarks. Table~\ref{tab:benchmarks} summarizes recent benchmarks in this field. In English, FINQA by \citet{chen-etal-2021-finqa} contains 8,281 question-answering pairs, with their numerical reasoning, from the earnings reports of S\&P 500 companies. ECTSum by~\cite{mukherjee-etal-2022-ectsum} contains 2,425 transcripts of earnings calls, paired with short telegram-style bullet point summaries. FLUE by~\cite{shah-etal-2022-flue} and FLARE by~\citet{xie-etal-2023-pixiu} offer heterogeneous benchmarks with various financial NLP tasks from existing datasets, including financial sentiment detection~\cite{malo-etal-2014-sentiment}, named entity recognition~\cite{alvarado-etal-2015-domain}, news headline classification~\cite{sinha-and-khandait-2020-impact}, question answering~\cite{maia-etal-2018-finqa}, boundary detection tasks~\cite{au-etal-2021-finsbd}, text summarization~\cite{zhou-etal-2021-trade,mukherjee-etal-2022-ectsum}, and stock movement prediction~\cite{xu-cohen-2018-stock,wu-etal-2018-stock,soun-etal-2022-stock}. FinTextQA by \citet{chen-etal-2024-fintextqa} includes 1,262 long-form QA pairs from finance textbooks and government websites. BizBench by \citet{krumdick-etal-2024-bizbench} features eight quantitative reasoning tasks from professional exams, earnings reports, and other financial sources. FINANCEBENCH by~\citet{islam-etal-2023-financebench} includes 10,231 question-answer-evidence triplets from public filings and earnings reports.

In Chinese, the CCKS series has released several datasets specifically designed for various event extraction tasks~\cite{tianchi_2019_ccks_entity,tianchi_2020_ccks_entity,tianchi_2021_ccks_causal,tianchi_2022_ccks}. Additionally, significant effort has gone into creating evaluation datasets for document-level extraction from financial announcements, judgments, and news articles. These include DCFEE~\cite{yang-etal-2018-dcfee}, DuEE-Fin~\cite{han-etal-2022-dueefin}, Doc2EDAG~\cite{zheng-etal-2019-doc2edag}, and CFinDEE~\cite{zhang-etal-2024-cfindee}. Moreover, both BBT-CFLEB dataset~\cite{lu-etal-2023-bbtfin} and CFLUE dataset~\cite{zhu-etal-2024-cflue} serve as heterogeneous benchmarks that cover a wide range of NLP tasks, including multiple-choice question answering and reasoning, news and text classification, summarization, relation extraction, sentiment classification, question answering, and reading comprehension.

In addition to the benchmarks mentioned above, several datasets have been developed in the financial domain to support the training of financial LLMs, such as FLANG~\cite{shah-etal-2022-flue}, Pixiu~\cite{xie-etal-2023-pixiu}, InvestLM~\cite{yang-etal-2023-investlm}, FinGPT~\cite{yang-etal-2023-fingpt}, and DianJin-R1~\cite{zhu2025dianjin}. However, it is important to note that most of these datasets primarily rely on sources such as financial news, announcements, filings, and earnings reports. In contrast, our focus is on financial meetings, which provide unique insights and discussions that are often absent from existing benchmarks.

\subsection{Summarization and Question-Answering Benchmarks}
\noindent\paragraph{Summarization Benchmarks.} Recent studies~\cite{goyal-etal-2022-news,zhang-etal-2024-benchmarking,pu-etal-2023-summarization} indicate that human preferences strongly favor summaries produced by LLMs over those generated by fine-tuned models or even reference summaries. This highlights the need to create new datasets that can comprehensively evaluate the summarization abilities of LLMs. For instance, SumSurvey by~\citet{liu-etal-2024-sumsurvey} is a new dataset focused on summarizing long scientific survey papers. REFINESUMM by~\citet{patil-etal-2024-refinesumm} is tailored for image-text multimodal summarization. MovieSum by~\citet{saxena-keller-2024-moviesum} comprises 2,200 pairs of movie screenplays and their corresponding summaries. Additionally, HeSum by~\citet{paz-argaman-etal-2024-hesum} is a novel benchmark dataset specifically designed for the low-resource Hebrew language. Our research also emphasizes long context, but we concentrate on meetings as our primary source, allowing us to analyze communication dynamics within this particular area. Several meeting speech summarization benchmarks are available, such as AMI~\cite{carletta-etal-2005-ami} and ICSI~\cite{janin-etal-2003-icsi}, which consist of English-language video (or audio) recordings of meetings, typically spanning tens of hours.

\noindent\paragraph{Question-Answering Benchmarks.} Question Answering (QA) is a key task in NLP, with increasingly challenging benchmarks being developed, including those focused on the financial domain. For example, several benchmarks have been introduced to assess the long context understanding capabilities of LLMs~\cite{li-etal-2024-loogle,wang-etal-2024-loong,bai-etal-2024-longbench}. Additionally, many multimodal QA benchmarks have emerged to assess LLMs' capabilities in detecting and retrieving relevant information from various inputs, such as images, videos, texts, tables, and meetings~\cite{li-etal-2023-m3it,fu-etal-2023-mme,jin-etal-2024-mmtom,prasad-etal-2023-meetingqa}.

\section{\texttt{M$^3$FinMeeting}: A Financial Meeting Benchmark}
\label{sec:m3finmeeting}

\subsection{Overview}
\label{sec:overview}

A formal financial meeting typically involves a few participants and lasts one to two hours. It promotes discussions among key participants, allowing for decision-making and strategic planning based on real-time information and reports. Participants express their opinions verbally, which distinguishes both the content and style of financial meetings from financial news, earnings reports, or announcements. The \texttt{M$^3$FinMeeting} benchmark, based on hundreds real financial meetings, includes three NLP tasks: summarization, question answering, and QA pair extraction. To reflect real-world scenarios, we gather financial meetings from various sectors defined by GICS: Communication Services, Consumer Discretionary, Consumer Staples, Energy, Financials, Healthcare, Industrials, Information Technology (IT), Materials, Real Estate, and Utilities. Additionally, all audio recordings of the meetings are transcribed using a state-of-the-art automatic speech recognition (ASR) toolkit, followed by manual corrections. In total, \texttt{M$^3$FinMeeting} includes 100 meetings in English (EN), 400 in Chinese (ZH), and 100 in Japanese (JA). Each meeting lasts an average of one hour. We employ the tiktoken tokenizer\footnote{\url{https://platform.openai.com/tokenizer} (cl100k\_base)} to process all transcriptions. Table~\ref{tab:data_stat} presents detailed data statistics. \ignore{Although the average duration of meetings is similar across the three languages, the average token count is highest for Chinese, followed by Japanese, and then English. This is because the tokenizer tends to break down traditional Chinese and Japanese words into multiple subwords, leading to a higher token count for these languages compared to English. The following sections will provide a detailed description of the evaluation task and benchmark construction.}

\begin{table}[!t]
\centering
\small
\begin{tabular}{l|l|l|l}
\toprule
\rowcolor{Gray}
\multicolumn{4}{c}{\bf Overview} \\
\toprule
\bf Language & \bf \#Meeting & \bf Avg hour &\bf Avg token \\
\midrule
EN & 100 & 0.96 & 10,086 \\
ZH & 400 & 1.15 & 11,740 \\
JA & 100 & 1.01 & 13,284 \\
\bottomrule
\rowcolor{Gray}
\multicolumn{4}{c}{\bf GICS Sector} \\
\toprule
\bf Sector & \bf Language & \bf \#Meeting & \bf Avg token\\
\midrule
Com. Services & EN, ZH, JA & 36 & 11,240 \\
Con. Dis. & EN, ZH, JA & 122 & 11,104 \\
Con. Staples & EN, ZH, JA & 52 & 11,699 \\
Energy & EN, ZH, JA & 32 & 15,841 \\
Financials & EN, ZH, JA & 49 & 12,835 \\
Healthcare & EN, ZH, JA & 47 & 13,937 \\
Industrials & EN, ZH,  JA & 111 & 12,062 \\
IT & EN, ZH, JA & 98 & 13,430 \\
Materials & EN, ZH, JA & 32 & 11,393 \\
RealEstate & EN, ZH, JA & 13 & 15,270 \\
Utilities & EN, ZH, JA & 8 & 17,290 \\
\bottomrule
\rowcolor{Gray}
\multicolumn{4}{c}{\bf Length Set} \\
\toprule
\bf Length & \bf Language & \bf \#Meeting & \bf Avg token\\
\midrule
Set1 (0-5K) & EN, ZH, JA & 59 & 3,546 \\
Set2 (5-10K) & EN, ZH, JA & 164 & 7,509 \\
Set3 (10-15K) & EN, ZH, JA & 195 & 12,476 \\
Set4 (15-20K) & EN, ZH, JA & 124 & 17,419 \\
Set5 (>20K) & EN, ZH, JA & 58 & 25,281 \\
\bottomrule
\end{tabular}
\caption{Data statistics of \texttt{M$^3$FinMeeting} benchmark.}
\label{tab:data_stat}
\end{table}

\subsection{Evaluation Tasks}
Given that financial meetings typically last one to two hours, users often need a summary and answers to specific questions. To address this, we propose three tasks that closely align with real-world needs.

\subsubsection{Summarization}
The summarization task aims to evaluate LLMs' ability to efficiently condense lengthy speeches while preserving the main ideas. Typically, these transcribed speech documents can be structured into sections based on discussion topics, with each section having its own summary, which we refer to section summary. We concatenate these section summaries sequentially to create a summary of the entire document. LLMs must implicitly identify and segment the document into various sections and then extract key point from each. Given transcribed speech documents and their reference summaries, we follow~\citet{koh-etal-2022-survey} to compute the compression ratio of a source document length against its reference summary length at both token-level and sentence-level. As shown in Table~\ref{tab:summarization_stat}, on average an English meeting contains 9.20 section summaries totaling 927 tokens.

\begin{table}[!t]
\centering
\small
\begin{tabular}{l|l|l|l|l}
\toprule
\bf Language & \bf AST & \bf \# SS & \bf \texttt{C}$_{token}$ & \bf \texttt{C}$_{sent}$\\
\midrule
EN & 927 & 9.20 & 10.88 & 10.49 \\
ZH & 2,524 & 15.17 & 4.65 & 3.62 \\
JA & 1,149 & 8.24 & 11.56 & 11.92 \\
\bottomrule
\end{tabular}
\caption{Statistics for the summarization task. Here, AST is for average summary token, \#SS is for averaged number of section summary, \texttt{C}$_{token}$ and \texttt{C}$_{sent}$ for the compression ratio at token-level and sentence-level, respectively.}
\label{tab:summarization_stat}
\end{table}

\subsubsection{QA Pair Extraction}
The task of question-answer (QA) pair extraction involves identifying and extracting relevant QA pairs from transcribed financial meetings. This is crucial for analyzing discussions and making key insights readily accessible. To successfully perform this task, LLMs must recognize various types of questions posed during the meeting and accurately locate their corresponding answers. For example, questions like \textit{What were we just talking about?} should be disregarded as they lack meaningful information. Additionally, participants may ask multiple questions at once, while responses may address them sequentially. This complexity requires LLMs to discern the structure of the dialogue, correctly pair each question with its answer. By employing natural language processing techniques, this task enhances information retrieval efficiency and facilitates subsequent analyses, providing structured and contextualized insights from the meeting. Table~\ref{tab:qa_stat} shows statistics of QA pairs. 

\begin{table}[!t]
\centering
\small
\begin{tabular}{l|l|l|l|l}
\toprule
\bf Lang. & \bf \texttt{QA}$_{token}$ & \bf \#QA & \bf \texttt{Q}$_{token}$ & \bf \texttt{A}$_{token}$ \\
\midrule
EN & 2,239 & 17.23 & 17.62 & 110.19 \\
ZH & 2,852 & 16.10 & 36.44 & 148.95 \\
JA & 2,934 & 10.84 & 34.55 & 178.99 \\
\bottomrule
\end{tabular}
\caption{Statistics for both QA Pair Extraction and Question Answering Tasks. \texttt{QA}$_{token}$ is the average token length of all QA pairs per meeting, \#QA is the average number of QA pairs per meeting, and \texttt{Q}$_{token}$ and \texttt{A}$_{token}$ denote the average token lengths of a question and answer, respectively.}
\label{tab:qa_stat}
\end{table}

\ignore{
\begin{table}[!t]
\centering
\small
\begin{tabular}{l|l|l|l|l|l}
\toprule
\bf Lang. & \texttt{QA}$_{token}$ & \texttt{Total}$_{QA}$ & \texttt{QA}$_{nunms}$ & \texttt{Q}$_{token}$ & \texttt{A}$_{token}$ \\
\midrule
EN & 2,239 & 1,723 & 17.46 & 39.51 & 161.46 \\
ZH & 2,852 & 6,442 & 17.40 & 17.80 & 111.30 \\
JA & 2,909 & 1,084 & 14.26 & 45.53 & 235.75 \\
\bottomrule
\end{tabular}
\caption{Statistics for both QA pair extraction and question answering tasks. Here, \texttt{QA}$_{token}$ indicates the average number of tokens per meeting, \texttt{QA}$_{nunms}$ is the average number of QA pairs per meeting, while \texttt{Q}$_{token}$ and \texttt{A}$_{token}$ signify the average token lengths for questions and answers, respectively.}
\label{tab:qa_stat}
\end{table}
}

\subsubsection{Question Answering}
The question answering (QA) task evaluates the ability of LLMs to localize knowledge, which is essential for effective long-context processing~\cite{wang-etal-2024-loong}. For simplicity, we use the QA pairs described above for this task. As mentioned, the transcribed speech text can be divided into multiple sections, the QA task tests the LLMs' capability to find evidence within that designated section, while other sections with similar but unrelated content act as noise. This setup ensures a focused assessment of the models' information retrieval skills. 

\ignore{
\begin{figure}
\centering
\resizebox{\columnwidth}{!}{
\includegraphics[width=0.9\linewidth, trim=26 0 7 0,clip]{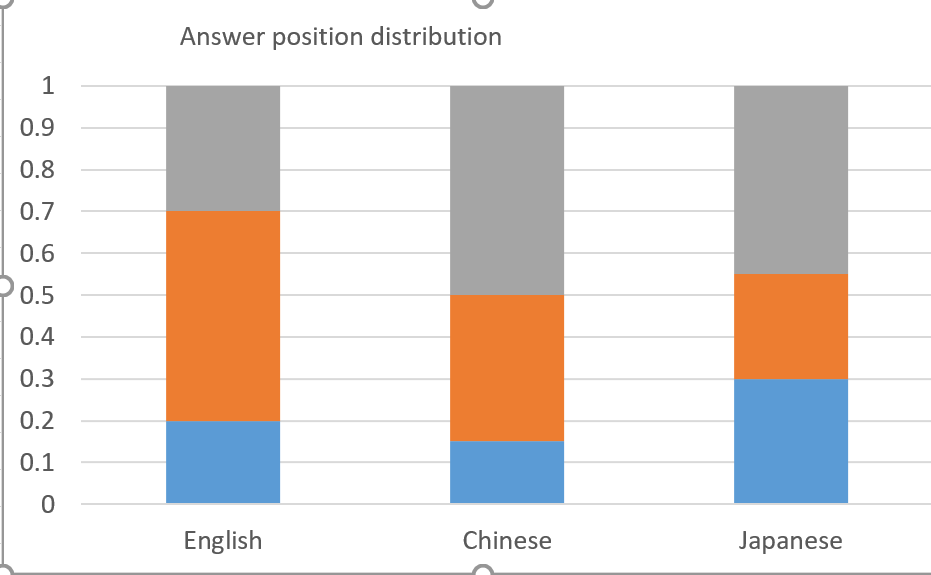}
}
\caption{Distribution of answer positions within the document.}
\label{fig:distribution}
\end{figure}}

\subsection{Benchmark Construction}
\subsubsection{Data Collection}
We have established four criteria for the manual collection of audio files from financial meetings, covering a range of events such as public roadshows, brokerage strategy meetings, industry exchanges, and earnings presentations: (1) Timeliness: Most meetings should be from recent years; (2) Length: Preference is given to longer audio files; (3) Categorizability: The audio files must align with categories defined in the GICS; (4) Authoritativeness: All audio files are sourced from our financial firm partners and are protected by our copyright. 

All audio files are transcribed into text using ASR toolkit Whisper,\footnote{\url{https://openai.com/index/whisper/}} followed by a thorough manual correction process. Strict measures are used to ensure that no sensitive or personally identifiable information is included in the transcripts. 

\subsubsection{Annotation Process and Quality Control}
The datasets for manual annotation are drawn from various projects that recruit experienced analysts fluent in different languages. These annotators follow the annotation guidelines (See Figure~\ref{fig:annotation_guideline} in Appendix~\ref{apx:annotation_guideline}), undergo comprehensive training, and have access to onboarding and guidance materials. Additionally, other analysts review the annotators' work and provide ongoing feedback\ignore{ throughout the process}.

\begin{itemize} 
\item Transcribed Speech Correction: Annotators correct ASR-generated transcriptions with original audio files available for reference during the correction process.
\item Summarization: The annotators segment each document into sections based on distinct topics, ensuring that only sections with clear boundaries are selected for summarization. For each valid section, annotators are encouraged to use simple sentences in the summary. Prior to the annotation process, annotators undergo extensive training and discussions to achieve a high level of consensus. In the annotation process, the original audio files are available for reference. \ignore{Each language project spans several weeks, reflecting the time-consuming nature of summarization.}
\item QA Pair Extraction: With corrected speech documents, annotators manually extract financially relevant questions from the text. For each identified question, they search for corresponding answers in the subsequent content. Only questions with valid answers are retained for further analysis. This meticulous process guarantees the quality and relevance of the extracted QA pairs, significantly enhancing the dataset's value for financial insights. 
\end{itemize}

Figure~\ref{fig:example} in Appendix~\ref{apx:example} shows a screenshot of an annotated example. For additional complete examples, please refer to the attached file.

\section{Experimentation}
\subsection{Experimental Settings}

\noindent\paragraph{Models.} \ignore{To provide a comprehensive assessment of the capabilities of LLMs in the context of financial meetings, w}We evaluate seven advanced long-context LLMs with context windows ranging from 16K to 1000K, including two API-based LLMs: GPT-4o-2024-08-06-128K~\cite{openai-2023-gpt4} and GPT-3.5.turbo-0125-16K, as well as five open-source LLMs: GLM4-9B-Chat-1000K~\cite{zeng-etal-2022-glm}, Llama3.1-8b-Instruct-128K~\cite{dubey-etal-2024-llama3}, Qwen2-7B-chat-128K~\cite{yang-etal-2024-qwen2}, Qwen2-72B-Instruct-128K, and Qwen2.5-72B-Instruct-128K~\cite{yang-etal-2024-qwen2.5}. All models support the languages in \texttt{M$^3$FinMeeting}.

\noindent\paragraph{Prompts.} We evaluate LLMs in a zero-shot setting. For summarization, we prompt the LLMs to implicitly identify document sections and generate individual summaries, which are then combined into a final document summary. For QA pair extraction, we first prompt the LLMs to extract all questions, then provide answers for each sequentially. For question answering, instead of addressing one question at a time, we combine related questions into a single prompt, allowing the LLM to produce a comprehensive response that includes all the answers. This better aligns with real-world tasks, like writing reviews or reports, and reduces API calls. Prompt examples are provided in Appendix~\ref{apx:prompt}.

\begin{figure}[t]
    \centering
    \includegraphics[width=0.8\columnwidth, trim={0cm 0cm 0cm 0cm}]{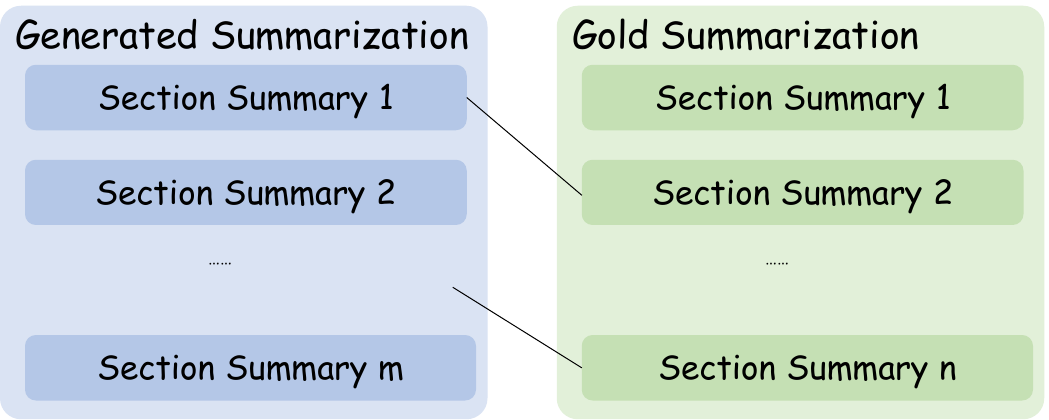}
    \caption{Illustration of the summarization evaluation.}
    \label{fig:summary_eval}
\end{figure}

\noindent\paragraph{Metrics.} For summarization, as shown in Figure~\ref{fig:summary_eval}, the final summary consists of multiple section summaries. We evaluate this using precision, recall, and F1 scores rather than traditional metrics like BLEU and ROUGE, as they better reflect how well the generated section summaries align with the reference (gold) summaries. Specifically, we align the automatic and gold section summaries using a cosine similarity score above 0.75, calculated with the OpenAI Embedding model\footnote{\url{https://platform.openai.com/docs/models/embeddings} (text-embedding-3-small)}. Appendix~\ref{apx:p_r_f1} gives details of the three metrics. Meanwhile, recent studies~\cite{wang-etal-2023-gpt4-eval, zhang-etal-2023-gpt4-eval, liu-etal-2024-gpt4-eval} show that GPT-4~\cite{openai-2023-gpt4} aligns closely with human evaluations. Therefore, we use GPT-4 (gpt-4-turbo-2024-04-09) as the judge (GPT-4-Judge) to evaluate document-level summaries based on five criteria: Coverage, Redundancy, Readability, Accuracy, and Consistency, with scores ranging from 0 to 100. We report the average GPT-4-Judge score, based on the prompt in Appendix~\ref{apx:prompt}. 

For QA pair extraction, we assess the quality of generated questions using precision, recall, and F1 scores, comparing them to the reference questions, similar to the summarization evaluation. Additionally, GPT-4-Judge evaluates the generated QA pairs against the gold pairs using the same five criteria, and we report the average score.

For question answering, we group all questions together and prompt the LLMs to answer them sequentially, repeating each question before generating its corresponding answer. Therefore, we evaluate performance using the same metrics as in QA pair extraction, assessing both the quality of repeated questions and the overall performance of the generated QA pairs.

For the three tasks, Appendix~\ref{apx:performance_bleu_rouge} presents performance metrics using BLEU~\cite{papineni-etal-2002-bleu} and ROUGE~\cite{lin-hovy-2002-rouge}. Appendix~\ref{apx:detailed_performance} includes detailed performance across languages, lengths, and GICS sectors. To address potential self-bias in LLM evaluations (e.g., GPT-4-Judge favoring GPT-4-generated answers), we follow \citet{bai-etal-2024-mtbench} and use Qwen-plus\footnote{\url{https://help.aliyun.com/zh/model-studio/developer-reference/what-is-qwen-llm}} as an alternative judge model. Results, shown in Section ~\ref{performance_qwen_plus}, indicate minimal bias, as the performance trend from GPT-4-Judge and Qwen-plus-Judge are very consistent. Moreover, we perform a human evaluation and calculate Fleiss' Kappa~\cite{scott-95-kappa} to measure agreement between GPT-4-Judge and human annotators, further supporting our conclusions. The results are presented in Section ~\ref{human_eval}. 

\subsection{Results}

\noindent\paragraph{Main Results.} 

\begin{table*}[!t]
\centering
\resizebox{\textwidth}{!}{
\begin{tabular}{l|llll|llll|llll|l}
\toprule
\multirow{2}{*}{\bf Model} & \multicolumn{4}{c|}{\bf Summarization} & \multicolumn{4}{c|}{\bf QA Pair Extraction} & \multicolumn{4}{c|}{\bf Question Answering} & \bf Overall\\
\cmidrule{2-14}
& \bf P. & \bf R. & \bf F1 & \bf GPT-4 & \bf P. & \bf R. & \bf F1 & \bf GPT-4 & \bf P. & \bf R. & \bf F1 & \bf GPT-4 & \bf GPT-4\\
\midrule
GPT-4o & 27.82 & 12.07 & 16.83 & 73.61 & \underline{23.33} & \underline{41.98} & \underline{29.99} & \underline{66.85} & \bf 93.93 & 93.16 & \underline{93.55} & 71.79 & \underline{70.66}  \\
\rowcolor{Gray}
GPT-3.5-turbo & 17.13 & 9.39 & 12.13 & 44.56 & 9.66 & 25.07 & 13.95 & 31.13 & 84.16 & 92.60 & 88.18 & 42.78 & 39.55 \\
GLM4-9B-Chat & 10.30 & 11.26 & 10.76 & 67.71 & 15.08 & 5.22 & 7.76 & 46.06 & 93.37 & 92.20 & 92.78 & 67.72 & 60.76 \\
\rowcolor{Gray}
LLaMA3.1-8B-Instruct & 6.24 & 8.34 & 7.14 & 52.01 & 13.05 & 7.30 & 9.37 & 44.64 & 57.21 & 39.97 & 47.06 & 40.01 & 45.76 \\
Qwen2-7B-Instruct & 28.67 & \underline{19.39} & \underline{23.14} & 73.59 & 11.98 & 16.40 & 13.85 & 37.33 & 89.83 & 93.01 & 91.40 & 69.99 & 60.71 \\
\rowcolor{Gray}
Qwen2-72B-Instruct & \bf 29.59 & \bf 20.18 & \bf 23.99 & \underline{74.17} & 22.43 & 28.82 & 25.22 & 60.85 & 93.27 & \underline{93.58} & 93.42 & \underline{73.50} & 69.66 \\
Qwen2.5-72B-Instruct & \underline{28.98} & 15.56 & 20.25 & \bf 74.51 & \bf 32.61 & \bf 45.65 & \bf 38.41 & \bf 68.03 & \underline{93.75} & \bf 93.59 & \bf 93.66 & \bf 74.81 & \bf 72.54 \\
\bottomrule
\end{tabular}
}
\caption{Performance of LLMs on three evaluation tasks. The overall GPT-4-Judge score is the micro-average of its scores across all three tasks. Scores in \textbf{bold}/\underline{underline} denote the top/second-best performances.}
\label{tab:main_result}
\end{table*}

\begin{figure*}[!t]
\centering
\includegraphics[width=\textwidth]{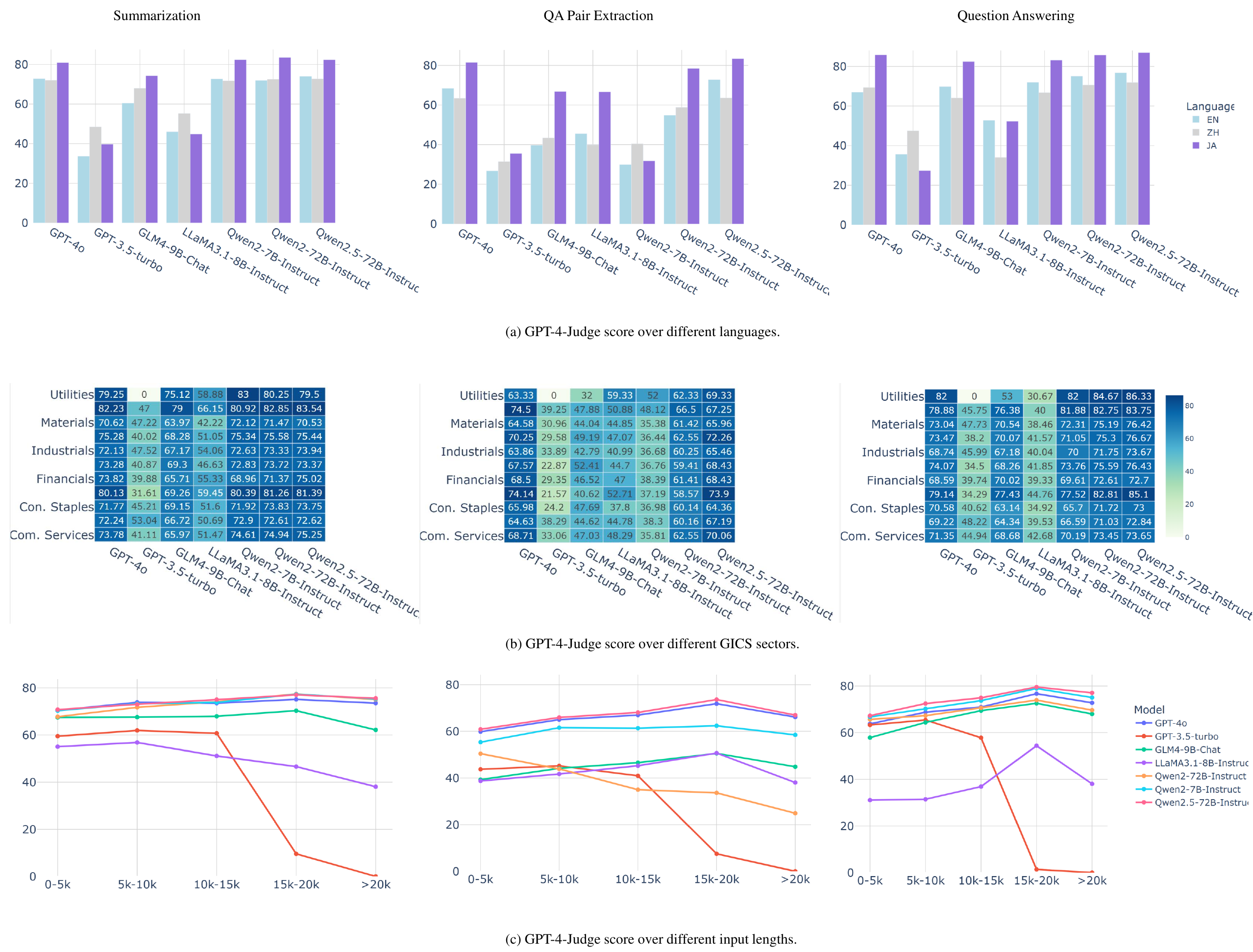}
\caption{Performance based on GPT-4-Judge scores across languages (a), GICS sectors (b), and input lengths (c).}
\label{fig:performance}
\end{figure*}

Table~\ref{tab:main_result} shows the main results over all meetings.\footnote{The total cost for OpenAI API calls was about \$2,500, while experiments with other LLMs use eight NVIDIA A100/80G GPUs.} From it, we have the following observations:
\begin{itemize}
\item \textbf{Overall performance}: The seven LLMs can be divided into three groups. Group 1 consists of Qwen2.5-72B-Instruct, Qwen2-72B-Instruct, and GPT-4o, all achieving an overall GPT-4-Judge score near or above 70.0. Among these, Qwen2.5-72B-Instruct performs best, followed by GPT-4o and Qwen2-72B-Instruct, which deliver comparable results. Group 2 includes Qwen2-7B-Instruct and GLM4-9B-Chat, both scoring around 60.0. Group 3 consists of GPT-3.5-turbo and LLaMA3.1-8B-Instruct, with LLaMA3.1-8B-Instruct outperforming GPT-3.5-turbo. 
\item \textbf{Summarization}: The precision, recall, and F1 scores for section-level summaries are all below 30\%, indicating poor alignment between generated and gold section-level summaries. These low scores suggest that LLMs struggle both with semantic accuracy and document segmentation. 
\item \textbf{QA Pair Extraction}: The low precision, recall, and F1 scores suggest poor alignment between generated and gold questions. For example, even the best-performing LLM, Qwen2.5-72B-Instruct, achieves only 45.65\% recall, missing more than half of the gold questions. This highlights significant room for improvement in extracting relevant QA pairs. 
\item \textbf{Question Answering}: The performance of all LLMs—measured by precision, recall, F1, and GPT-4-Judge scores—is significantly higher than for QA pair extraction.\footnote{One exception is LLaMA3.1-8B-Instruct, which has difficulty following instructions and often fails to repeat the questions, resulting in lower GPT-4-Judge scores for the question answering task.} This discrepancy is not surprising, as in the question answering task, the questions are explicitly provided in the prompt. High F1 scores (over 90\%) show that most LLMs can follow instructions well and properly repeat questions. 
\end{itemize}

\noindent\paragraph{Effect over Different Languages.} 
Figure~\ref{fig:performance} (a) shows the GPT-4-Judge scores across three languages. Most models perform best in Japanese, but there is no clear advantage in either Chinese or English. A closer look reveals that LLMs are more consistent in Japanese, likely because they adhere to the instructions more effectively in this language. The three Qwen models perform similarly in the Summarization task across all languages, followed by the Question Answering task. However, their performance varies most in the QA pair extraction task, where Qwen2.5-72B-Instruct obtains the highest scores, outperforming Qwen2-7B-Instruct with GPT-4-Judge scores of 72.76, 63.59, and 83.45 for English, Chinese, and Japanese, respectively. 

\noindent\paragraph{Effect over Different Sectors.} 
Figure~\ref{fig:performance} (b) compares model performance across sectors. Communication Services, Consumer Discretionary, and IT generally achieve higher GPT-4-Judge scores in summarization and question answering. However, the performance trend becomes more complex for the QA pair extraction task, showing increased variability across sectors. Overall, the performance gaps among sectors for GPT-4o, Qwen2-72B-Instruct, and Qwen2.5-72B-Instruct are much smaller compared to GPT-3.5-turbo and LLaMA3.1-8B-Instruct.\footnote{GPT-3.5-turbo fails for the Utilities sector due to input length exceeding its 16K token limit.} This suggests that the former models are less affected by sector differences.

\noindent\paragraph{Effect over Different Input Lengths.}
Figure~\ref{fig:performance} (c) compares the performance across varying input lengths. A key observation is the sharp drop in GPT-3.5-turbo’s performance when the input exceeds 15K tokens, due to its 16K token context limit. In contrast, both Qwen2.5-72B-Instruct and GPT-4o demonstrate stable and competent performance across the three tasks, particularly excelling in handling longer contexts exceeding 15K tokens. On the other hand, Qwen2-72B-Instruct shows a declining trend in the QA pair extraction task as input length increases, indicating a reduced capability to maintain performance with longer inputs in this specific task. Future research could explore structured modeling, as outlined by ~\cite{zhu-etal-2019-modeling}, to improve handling of long input contexts.

\begin{figure}[t]
\centering
\includegraphics[width=1.0\linewidth]{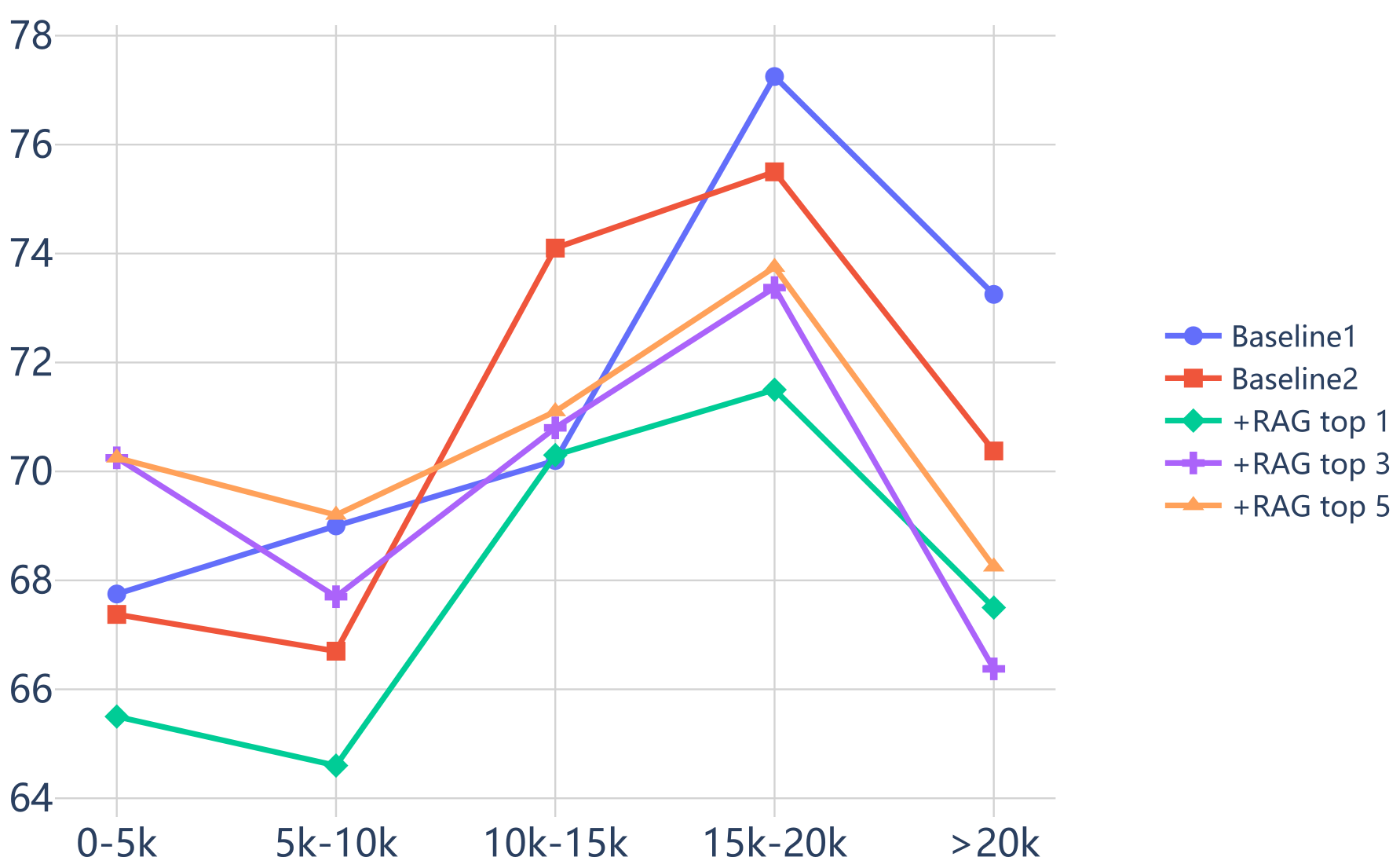}
\caption{Performance (in GPT-4-Judge score) across different input lengths before or after adding RAG module. Baseline 1 refers to prompting the LLM to answer a list of questions, while Baseline 2 refers to answering one question per response.}
\label{fig:rag}
\end{figure}

\noindent\paragraph{Effect of RAG-based Question Answering.}
Instead of prompting the LLMs to answer a list of questions in a single response, we also explore RAG-based question answering, where the LLM answers questions individually based on retrieved document chunks. Following \citet{wang-etal-2024-loong}, we divide the document into 1,024-token chunks. For embedding, we utilize the OpenAI Embedding model. Figure~\ref{fig:rag} compares the performance of Qwen2.5-72B-Instruct before and after the integration of the RAG module. This comparison is based on a random selection of 10 meetings from each length set, resulting in a total of 50 meetings. The results indicate that for documents exceeding 15K tokens, answering all questions in a single response (Baseline 1) outperforms all other variants that answer questions one at a time. Additionally, for variants that respond to one question at a time in documents longer than 10K tokens, we observe that a larger context leads to better performance, specifically: Baseline 2 > RAG (top 5) > RAG (top 3) > RAG (top 1). Notably, RAG (with top 5) only surpasses the non-RAG variants for documents shorter than 10K tokens.

\subsection{Performance Evaluated by Qwen-plus-Judge}
\label{performance_qwen_plus}

\begin{table*}[!th]
\centering
\small
\begin{tabular}{l|ccc|c}
\toprule
\bf Model & \bf Summarization & \bf QA Pair Extraction & \bf Question Answering & \bf Overall \\
\hline
GPT-4o & 76.01 & \underline{67.02} & 69.84 & \underline{71.12}\\
\rowcolor{Gray}
GPT-3.5-turbo & 45.46  & 31.43 & 37.88 & 38.48\\
GLM4-9B-Chat & 68.20  & 47.06 & 60.39 & 58.86\\
\rowcolor{Gray}
LLaMA3.1-8B-Instruct & 51.46  & 43.90 & 32.05 & 42.75\\
Qwen2-7B-Instruct & 75.39 & 42.65 & 67.87 & 62.38\\
\rowcolor{Gray}
Qwen2-72B-Instruct & \underline{76.22}  & 65.58 & \underline{70.73} & 71.01\\
Qwen2.5-72B-Instruct & \bf 76.99  & \bf 70.25 & \bf 74.01 & \bf 73.85\\
\bottomrule
\end{tabular}
\caption{The results using Qwen-plus as the judge model.}
\label{tab:main_result_qwen}
\end{table*}

\begin{table*}[!th] 
\centering 
\small 
\begin{tabular}{l|ccc} 
\toprule 
\bf Evaluation Method & \bf Summarization & \bf QA Pair Extraction & \bf Question Answering \\ 
\hline 
GPT-4-Judge & 72.83 & 60.76 & 66.44 \\ 
Human Annotators & 3.68 & 3.13 & 3.36 \\ 
\bottomrule 
\end{tabular} 
\caption{Comparison of average performance ratings between GPT-4-Judge and human annotators. Note that GPT-4-Judge uses a 1-100 scale, while human annotators use a 1-5 scale.} 
\label{tab:comparison_gpt_human} 
\end{table*}

\begin{table}[!th] 
\centering 
\small 
\begin{tabular}{l|c} 
\toprule 
\bf Agreement & Kappa \\ 
\hline 
GPT-4-Judge \& Human Annotators & 0.701  \\ 
Human Annotators & 0.650  \\ 
\bottomrule 
\end{tabular} 
\caption{Fleiss’ Kappa score between GPT-4-Judge and human annotators.} 
\label{tab:kappa} 
\end{table}

In addition to the performance assessed by the GPT-4-Judge, as presented in Table~\ref{tab:main_result}, Table~\ref{tab:main_result_qwen} shows the performance evaluated by the Qwen-Plus-Judge. The prompt templates used for the Qwen-Plus-Judge are the same as those for GPT-4-Judge, as illustrated in Figure~\ref{fig:prompt_summarization_eval} and Figure~\ref{fig:prompt_qa_eval}. The performance trends in Table~\ref{tab:main_result_qwen} align closely with those in Table~\ref{tab:main_result}, where Qwen2.5-72B-Instruct remains the top-performing model, followed by GPT-4o and Qwen2-72B-Instruct, which show similar performance. Next are Qwen2-7B-Instruct and GLM4-9B-Chat, both of which also exhibit comparable performance. This consistency further demonstrates that Qwen-Plus-Judge is a reliable alternative evaluator.

\subsection{Human Evaluation and Fleiss' Kappa Agreement Between GPT-4-Judge and Human Evaluators}
\label{human_eval}

We randomly select 100 meetings and recruit five expert human annotators to assess the overall quality of GPT-4o's responses, including summarization, QA pair extraction, and question answering. Each response is rated on a scale from 1 to 5, based on the same five criteria used to prompt GPT-4-Judge. The final score for each response is the average rating across the five annotators. Table~\ref{tab:comparison_gpt_human} compares the performance assessed by GPT-4-Judge and human annotators.

Moreover, to assess the agreement among the six evaluators (five human annotators and GPT-4-Judge), we compute Fleiss' Kappa score, a metric for inter-annotator agreement. Specifically, we calculate two types of Kappa scores: 1) the agreement among all six evaluators, where GPT-4-Judge's ratings (on a 1-100 scale) are converted to a 1-5 scale by dividing by 20; and 2) the agreement among the five human annotators. As shown in Table~\ref{tab:kappa}, the agreement among GPT-4-Judge and human annotators is still higher than that among humans.
\ignore{
\begin{itemize}
\item Kappa score between GPT-4-Judge and human annotators: We first compute the Kappa score between GPT-4-Judge and each individual annotator, then report the average Kappa score.
\item Kappa score between human annotators: We calculate the Kappa score between each pair of human annotators and then report the average Kappa score. 
\end{itemize}
}

\subsection{Performance in BLEU and ROUGE} 

The analysis of model performances, based on BLEU and ROUGE metrics, highlights notable differences across various tasks. Specifically, Qwen2-72B-Instruct excels in the summarization task, consistently generating concise and coherent summaries that underscore its strength in synthesizing information. Meanwhile, Qwen2.5-72B-Instruct leads in both QA pair extraction and question answering, demonstrating its adeptness at understanding queries and providing precise responses. This comparison underscores the unique advantages and task-specific expertise of each model, offering a more comprehensive insight into their capabilities. For detailed data and specific scores, please refer to the full table located in the appendix~\ref{apx:performance_bleu_rouge}.

\section{Conclusion}
\label{sec:conclusion}

In this paper, we have introduced \texttt{M$^3$FinMeeting}, a novel multilingual, multi-sector, and multi-task benchmark specifically designed to evaluate financial meeting understanding in large language models (LLMs). By incorporating real-world dialogue from financial meetings, our dataset fills significant gaps in existing benchmarks, which often rely on static sources like news articles and earnings reports. Supporting English, Chinese, and Japanese, and encompassing 11 industry sectors defined by GICS, \texttt{M$^3$FinMeeting} enables a comprehensive evaluation of LLMs through tasks such as summarization, question-answer pair extraction, and question answering. Experimental results with seven representative LLMs, including GPT-4o and Qwen2.5-72B-Instruct, demonstrate notable performance challenges, highlighting significant room for improvement and establishing \texttt{M$^3$FinMeeting} as a valuable resource for advancing research in financial language processing.

\section*{Acknowledgments} The authors express their heartfelt gratitude to the anonymous reviewers for their invaluable feedback and insightful comments, which greatly enhanced the quality of this work. We also offer our sincere appreciation to the members of the Qwen DianJin Team for their exceptional contributions, dedication, and hard work, which were instrumental in the success of this project. This work was supported by the Alibaba Innovative Research Program.

\section*{Limitations}
Our work has several limitations. 
\begin{itemize}
\item High Annotation Costs and Challenges: Both summarization and question-answer pair extraction require annotators to summarize content and extract question-answer pairs from audio recordings lasting 1-2 hours and text exceeding 10K tokens. This process relies heavily on the annotators' professional expertise. Summarization, in particular, demands a high level of skill from professional analysts and often involves open-ended responses. Consequently, the annotation process necessitates a significant investment of time and effort.
\item Limited Coverage for Question Answering: In the question-answering task, we focus exclusively from the extracted QA pairs. This limits our ability to evaluate the LLMs' capacity to search for evidence within the document beyond the provided questions. As a result, the dataset may not fully capture the models' potential in more complex reasoning scenarios that require deeper comprehension of the content.
\item Evaluation Issues: The performance of summary alignment in summarization and question alignment in both question-answer pair extraction and question answering relies on the embeddings used. Due to budget constraints, we have not conducted a more extensive evaluation, which may affect the robustness of our findings. \ignore{Furthermore, we employ GPT-4 as the evaluation judge, which could introduce potential biases when assessing different LLMs.}
\end{itemize}

\section*{Ethics}
Our dataset is sourced from publicly mandated disclosures, such as earnings calls and roadshows. While this information is publicly available, we anonymize it to respect corporate preferences for dissemination. This ensures both the academic utility and ethical integrity of our benchmark. Specifically, we remove company identifiers and sensitive details using GPT-4, followed by manual verification, to maintain privacy without compromising research value. All data annotators are part of the funded projects, ensuring consistent and responsible data handling. The dataset, excluding original audio files, will be available online.

\bibliography{custom}

\begin{thebibliography}{60}
\providecommand{\natexlab}[1]{#1}

\bibitem[{Alvarado et~al.(2015)Alvarado, Verspoor, and Baldwin}]{alvarado-etal-2015-domain}
Julio Cesar~Salinas Alvarado, Karin Verspoor, and Timothy Baldwin. 2015.
\newblock Domain adaption of named entity recognition to support credit risk assessment.
\newblock In \emph{Proceedings of the Australasian Language Technology Association Workshop}, pages 84--90.

\bibitem[{Au et~al.(2021)Au, Ait-Azzi, and Kang}]{au-etal-2021-finsbd}
Willy Au, Abderrahim Ait-Azzi, and Juyeon Kang. 2021.
\newblock Finsbd-2021: The 3rd shared task on structure boundary detection in unstructured text in the financial domain.
\newblock In \emph{Companion Proceedings of WWW}, page 276–279.

\bibitem[{Bai et~al.(2024{\natexlab{a}})Bai, Liu, Bu, He, Liu, Zhou, Lin, Su, Ge, Zheng, and Ouyang}]{bai-etal-2024-mtbench}
Ge~Bai, Jie Liu, Xingyuan Bu, Yancheng He, Jiaheng Liu, Zhanhui Zhou, Zhuoran Lin, Wenbo Su, Tiezheng Ge, Bo~Zheng, and Wanli Ouyang. 2024{\natexlab{a}}.
\newblock {MT}-bench-101: A fine-grained benchmark for evaluating large language models in multi-turn dialogues.
\newblock In \emph{Proceedings of ACL}, pages 7421--7454.

\bibitem[{Bai et~al.(2024{\natexlab{b}})Bai, Lv, Zhang, Lyu, Tang, Huang, Du, Liu, Zeng, Hou, Dong, Tang, and Li}]{bai-etal-2024-longbench}
Yushi Bai, Xin Lv, Jiajie Zhang, Hongchang Lyu, Jiankai Tang, Zhidian Huang, Zhengxiao Du, Xiao Liu, Aohan Zeng, Lei Hou, Yuxiao Dong, Jie Tang, and Juanzi Li. 2024{\natexlab{b}}.
\newblock {L}ong{B}ench: A bilingual, multitask benchmark for long context understanding.
\newblock In \emph{Proceedings of ACL}, pages 3119--3137.

\bibitem[{Carletta et~al.(2005)Carletta, Ashby, Bourban, Flynn, Guillemot, Hain, Kadlec, Karaiskos, Kraaij, Kronenthal, Lathoud, Lincoln, Lisowska, McCowan, Post, Reidsma, and Wellner}]{carletta-etal-2005-ami}
Jean Carletta, Simone Ashby, Sebastien Bourban, Mike Flynn, Mael Guillemot, Thomas Hain, Jaroslav Kadlec, Vasilis Karaiskos, Wessel Kraaij, Melissa Kronenthal, Guillaume Lathoud, Mike Lincoln, Agnes Lisowska, Iain McCowan, Wilfried Post, Dennis Reidsma, and Pierre Wellner. 2005.
\newblock The ami meeting corpus: a pre-announcement.
\newblock In \emph{Proceedings of MLMI}, page 28–39.

\bibitem[{Chen et~al.(2024)Chen, Zhou, Hua, Xin, Chen, Li, Zhu, and Liang}]{chen-etal-2024-fintextqa}
Jian Chen, Peilin Zhou, Yining Hua, Loh Xin, Kehui Chen, Ziyuan Li, Bing Zhu, and Junwei Liang. 2024.
\newblock {F}in{T}ext{QA}: A dataset for long-form financial question answering.
\newblock In \emph{Proceedings of ACL}, pages 6025--6047.

\bibitem[{Chen et~al.(2021)Chen, Chen, Smiley, Shah, Borova, Langdon, Moussa, Beane, Huang, Routledge, and Wang}]{chen-etal-2021-finqa}
Zhiyu Chen, Wenhu Chen, Charese Smiley, Sameena Shah, Iana Borova, Dylan Langdon, Reema Moussa, Matt Beane, Ting-Hao Huang, Bryan Routledge, and William~Yang Wang. 2021.
\newblock {F}in{QA}: A dataset of numerical reasoning over financial data.
\newblock In \emph{Proceedings of EMNLP}, pages 3697--3711.

\bibitem[{Chen et~al.(2022)Chen, Li, Smiley, Ma, Shah, and Wang}]{chen-etal-2022-convfinqa}
Zhiyu Chen, Shiyang Li, Charese Smiley, Zhiqiang Ma, Sameena Shah, and William~Yang Wang. 2022.
\newblock {C}onv{F}in{QA}: Exploring the chain of numerical reasoning in conversational finance question answering.
\newblock In \emph{Proceedings of EMNLP}, pages 6279--6292.

\bibitem[{Dubey et~al.(2024)Dubey, Jauhri, Pandey, Kadian, Al-Dahle, Letman, Mathur, Schelten, Yang, Fan et~al.}]{dubey-etal-2024-llama3}
Abhimanyu Dubey, Abhinav Jauhri, Abhinav Pandey, Abhishek Kadian, Ahmad Al-Dahle, Aiesha Letman, Akhil Mathur, Alan Schelten, Amy Yang, Angela Fan, et~al. 2024.
\newblock The llama 3 herd of models.
\newblock \emph{Computing Research Repository}, arXiv:2407.21783.

\bibitem[{Fu et~al.(2023)Fu, Chen, Shen, Qin, Zhang, Lin, Yang, Zheng, Li, Sun et~al.}]{fu-etal-2023-mme}
Chaoyou Fu, Peixian Chen, Yunhang Shen, Yulei Qin, Mengdan Zhang, Xu~Lin, Jinrui Yang, Xiawu Zheng, Ke~Li, Xing Sun, et~al. 2023.
\newblock Mme: A comprehensive evaluation benchmark for multimodal large language models.
\newblock \emph{Computing Research Repository}, arXiv:2306.1339.

\bibitem[{Goyal et~al.(2022)Goyal, Li, and Durrett}]{goyal-etal-2022-news}
Tanya Goyal, Junyi~Jessy Li, and Greg Durrett. 2022.
\newblock News summarization and evaluation in the era of gpt-3.
\newblock \emph{Computing Research Repository}, arXiv:2209.12356.

\bibitem[{Han et~al.(2022)Han, Zhang, Li, Xu, Peng, and Zeng}]{han-etal-2022-dueefin}
Cuiyun Han, Jinchuan Zhang, Xinyu Li, Guojin Xu, Weihua Peng, and Zengfeng Zeng. 2022.
\newblock Duee-fin: A large-scale dataset for document-level event extraction.
\newblock In \emph{Proceedings of NLPCC}, pages 172--183.

\bibitem[{Islam et~al.(2023)Islam, Kannappan, Kiela, Qian, Scherrer, and Vidgen}]{islam-etal-2023-financebench}
Pranab Islam, Anand Kannappan, Douwe Kiela, Rebecca Qian, Nino Scherrer, and Bertie Vidgen. 2023.
\newblock Financebench: A new benchmark for financial question answering.
\newblock \emph{Computing Research Repository}, arXiv:2311.11944.

\bibitem[{Janin et~al.(2003)Janin, Baaron, Edwards, Ellis, Gelbart, Morgan, Peskin, Pfau, Shriberg, Stolcke, and Wooters}]{janin-etal-2003-icsi}
Adam Janin, Don Baaron, Jane Edwards, Dan Ellis, David Gelbart, Nelson Morgan, Barbara Peskin, Thilo Pfau, Elizabeth Shriberg, Andreas Stolcke, and Chunk Wooters. 2003.
\newblock The icsi meeting corpus.
\newblock In \emph{Proceedings of ICASSP}, pages 364--367.

\bibitem[{Jin et~al.(2024)Jin, Wu, Cao, Xiang, Kuo, Hu, Ullman, Torralba, Tenenbaum, and Shu}]{jin-etal-2024-mmtom}
Chuanyang Jin, Yutong Wu, Jing Cao, Jiannan Xiang, Yen-Ling Kuo, Zhiting Hu, Tomer Ullman, Antonio Torralba, Joshua Tenenbaum, and Tianmin Shu. 2024.
\newblock {MMT}o{M}-{QA}: Multimodal theory of mind question answering.
\newblock In \emph{Proceedings of ACL}, pages 16077--16102.

\bibitem[{Koh et~al.(2022)Koh, Ju, Liu, and Pan}]{koh-etal-2022-survey}
Huan~Yee Koh, Jiaxin Ju, Ming Liu, and Shirui Pan. 2022.
\newblock An empirical survey on long document summarization: Datasets, models, and metrics.
\newblock \emph{ACM Comput. Surv.}, 55(8).

\bibitem[{Krumdick et~al.(2024)Krumdick, Koncel-Kedziorski, Lai, Reddy, Lovering, and Tanner}]{krumdick-etal-2024-bizbench}
Michael Krumdick, Rik Koncel-Kedziorski, Viet Lai, Varshini Reddy, Charles Lovering, and Chris Tanner. 2024.
\newblock {B}iz{B}ench: A quantitative reasoning benchmark for business and finance.
\newblock In \emph{Proceedings of ACL}, pages 8309--8332.

\bibitem[{Li et~al.(2024)Li, Wang, Zheng, and Zhang}]{li-etal-2024-loogle}
Jiaqi Li, Mengmeng Wang, Zilong Zheng, and Muhan Zhang. 2024.
\newblock {L}oo{GLE}: Can long-context language models understand long contexts?
\newblock In \emph{Proceedings of ACL}, pages 16304--16333.

\bibitem[{Li et~al.(2023)Li, Yin, Li, Chen, Wang, Ren, Li, Yang, Xu, Sun, Kong, and Liu}]{li-etal-2023-m3it}
Lei Li, Yuwei Yin, Shicheng Li, Liang Chen, Peiyi Wang, Shuhuai Ren, Mukai Li, Yazheng Yang, Jingjing Xu, Xu~Sun, Lingpeng Kong, and Qi~Liu. 2023.
\newblock M$^3$it: A large-scale dataset towards multi-modal multilingual instruction tuning.
\newblock \emph{Computing Research Repository}, arXiv:2306.04387.

\bibitem[{Lin and Hovy(2002)}]{lin-hovy-2002-rouge}
Chin-Yew Lin and Eduard Hovy. 2002.
\newblock Automatic evaluation of summaries using n-gram co-occurrence statistics.
\newblock In \emph{Proceedings of ACL}, page 311–318.

\bibitem[{Liu et~al.(2024{\natexlab{a}})Liu, Liu, Yu, Zhang, Jiang, Li, and Huang}]{liu-etal-2024-sumsurvey}
Ran Liu, Ming Liu, Min Yu, He~Zhang, Jianguo Jiang, Gang Li, and Weiqing Huang. 2024{\natexlab{a}}.
\newblock {S}um{S}urvey: An abstractive dataset of scientific survey papers for long document summarization.
\newblock In \emph{Findings of ACL}, pages 9632--9651.

\bibitem[{Liu et~al.(2024{\natexlab{b}})Liu, Yang, Huang, Zhang, Huang, Wei, Deng, Sun, and Zhang}]{liu-etal-2024-gpt4-eval}
Yuxuan Liu, Tianchi Yang, Shaohan Huang, Zihan Zhang, Haizhen Huang, Furu Wei, Weiwei Deng, Feng Sun, and Qi~Zhang. 2024{\natexlab{b}}.
\newblock Calibrating llm-based evaluator.
\newblock In \emph{Proceedings of LREC-COLING}, pages 2638--2656.

\bibitem[{Lu et~al.(2023)Lu, Wu, Liang, Xu, He, Geng, Han, Xin, and Xiao}]{lu-etal-2023-bbtfin}
Dakuan Lu, Hengkui Wu, Jiaqing Liang, Yipei Xu, Qianyu He, Yipeng Geng, Mengkun Han, Yingsi Xin, and Yanghua Xiao. 2023.
\newblock Bbt-fin: Comprehensive construction of chinese financial domain pre-trained language model, corpus and benchmark.
\newblock \emph{Computing Research Repository}, arXiv:2302.09432.

\bibitem[{Maia et~al.(2018)Maia, Handschuh, Freitas, Davis, McDermott, Zarrouk, and Balahur}]{maia-etal-2018-finqa}
Macedo Maia, Siegfried Handschuh, Andr\'{e} Freitas, Brian Davis, Ross McDermott, Manel Zarrouk, and Alexandra Balahur. 2018.
\newblock Www'18 open challenge: Financial opinion mining and question answering.
\newblock In \emph{Companion Proceedings of WWW}, page 1941–1942.

\bibitem[{Malo et~al.(2014)Malo, Sinha, Takala, Korhonen, and Wallenius}]{malo-etal-2014-sentiment}
Pekka Malo, Ankur Sinha, Pyry Takala, Pekka Korhonen, and Jyrki Wallenius. 2014.
\newblock Good debt or bad debt: Detecting semantic orientations in economic texts.
\newblock \emph{Journal of the American Society for Information Science and Technology}, 65:782--796.

\bibitem[{Mukherjee et~al.(2022)Mukherjee, Bohra, Banerjee, Sharma, Hegde, Shaikh, Shrivastava, Dasgupta, Ganguly, Ghosh, and Goyal}]{mukherjee-etal-2022-ectsum}
Rajdeep Mukherjee, Abhinav Bohra, Akash Banerjee, Soumya Sharma, Manjunath Hegde, Afreen Shaikh, Shivani Shrivastava, Koustuv Dasgupta, Niloy Ganguly, Saptarshi Ghosh, and Pawan Goyal. 2022.
\newblock {ECTS}um: A new benchmark dataset for bullet point summarization of long earnings call transcripts.
\newblock In \emph{Proceedings of EMNLP}, pages 10893--10906.

\bibitem[{OpenAI(2023)}]{openai-2023-gpt4}
OpenAI. 2023.
\newblock Gpt-4 technical report.
\newblock \emph{Computing Research Repository}, arXiv:2303.08774.

\bibitem[{Papineni et~al.(2002)Papineni, Roukos, Ward, and Zhu}]{papineni-etal-2002-bleu}
Kishore Papineni, Salim Roukos, Todd Ward, and Wei-Jing Zhu. 2002.
\newblock Bleu: A method for automatic evaluation of machine translation.
\newblock In \emph{Proceedings of ACL}, page 311–318.

\bibitem[{Patil et~al.(2024)Patil, Ribeiro, Liu, Bansal, and Dreyer}]{patil-etal-2024-refinesumm}
Vaidehi Patil, Leonardo Ribeiro, Mengwen Liu, Mohit Bansal, and Markus Dreyer. 2024.
\newblock {REFINESUMM}: Self-refining {MLLM} for generating a multimodal summarization dataset.
\newblock In \emph{Proceedings of ACL}, pages 13773--13786.

\bibitem[{Paz-Argaman et~al.(2024)Paz-Argaman, Mondshine, Achi~Mordechai, and Tsarfaty}]{paz-argaman-etal-2024-hesum}
Tzuf Paz-Argaman, Itai Mondshine, Asaf Achi~Mordechai, and Reut Tsarfaty. 2024.
\newblock {H}e{S}um: a novel dataset for abstractive text summarization in {H}ebrew.
\newblock In \emph{Findings of ACL}, pages 6378--6388.

\bibitem[{Prasad et~al.(2023)Prasad, Bui, Yoon, Deilamsalehy, Dernoncourt, and Bansal}]{prasad-etal-2023-meetingqa}
Archiki Prasad, Trung Bui, Seunghyun Yoon, Hanieh Deilamsalehy, Franck Dernoncourt, and Mohit Bansal. 2023.
\newblock {M}eeting{QA}: Extractive question-answering on meeting transcripts.
\newblock In \emph{Proceedings of ACL}, pages 15000--15025.

\bibitem[{Pu et~al.(2023)Pu, Gao, and Wan}]{pu-etal-2023-summarization}
Xiao Pu, Mingqi Gao, and Xiaojun Wan. 2023.
\newblock Summarization is (almost) dead.
\newblock \emph{Computing Research Repository}, arXiv:2309.09558.

\bibitem[{Saxena and Keller(2024)}]{saxena-keller-2024-moviesum}
Rohit Saxena and Frank Keller. 2024.
\newblock {M}ovie{S}um: An abstractive summarization dataset for movie screenplays.
\newblock In \emph{Findings of ACL}, pages 4043--4050.

\bibitem[{Scott(1995)}]{scott-95-kappa}
William~A. Scott. 1995.
\newblock Reliability of content analysis: The case of nominal scale coding.
\newblock \emph{The Public Opinion Quarterly}, 19:321--325.

\bibitem[{Shah et~al.(2022)Shah, Chawla, Eidnani, Shah, Du, Chava, Raman, Smiley, Chen, and Yang}]{shah-etal-2022-flue}
Raj Shah, Kunal Chawla, Dheeraj Eidnani, Agam Shah, Wendi Du, Sudheer Chava, Natraj Raman, Charese Smiley, Jiaao Chen, and Diyi Yang. 2022.
\newblock When flue meets flang: Benchmarks and large pretrained language model for financial domain.
\newblock In \emph{Proceedings of EMNLP}, pages 2322--2335.

\bibitem[{Sinha and Khandait(2020)}]{sinha-and-khandait-2020-impact}
Ankur Sinha and Tanmay Khandait. 2020.
\newblock Impact of news on the commodity market: Dataset and results.
\newblock \emph{Computing Research Repository}, arXiv:2009.04202.

\bibitem[{Soun et~al.(2022)Soun, Yoo, Cho, Jeon, , and Kang}]{soun-etal-2022-stock}
Yejun Soun, Jaemin Yoo, Minyong Cho, Jihyeong Jeon, , and U~Kang. 2022.
\newblock Accurate stock movement prediction with self-supervised learning from sparse noisy tweets.
\newblock In \emph{Proceedings of CKIM}, pages 1691--1700.

\bibitem[{Tianchi(2019)}]{tianchi_2019_ccks_entity}
Tianchi. 2019.
\newblock Ccks2019 financial domain document-level event entity extraction dataset.
\newblock \url{https://tianchi.aliyun.com/dataset/111237}.

\bibitem[{Tianchi(2020)}]{tianchi_2020_ccks_entity}
Tianchi. 2020.
\newblock Ccks2020 financial domain document-level event entity extraction dataset.
\newblock \url{https://https://tianchi.aliyun.com/dataset/111209}.

\bibitem[{Tianchi(2021)}]{tianchi_2021_ccks_causal}
Tianchi. 2021.
\newblock Ccks2021 financial domain event causal relationship extraction dataset.
\newblock \url{https://tianchi.aliyun.com/dataset/110901}.

\bibitem[{Tianchi(2022)}]{tianchi_2022_ccks}
Tianchi. 2022.
\newblock Ccks2022 financial domain few-shot event extraction dataset.
\newblock \url{https://tianchi.aliyun.com/dataset/136800}.

\bibitem[{Wang et~al.(2023)Wang, Cheng, Guo, Yue, Ding, Xu, Wang, Hu, Zhang, and Zhang}]{wang-etal-2023-gpt4-eval}
Cunxiang Wang, Sirui Cheng, Qipeng Guo, Yuanhao Yue, Bowen Ding, Zhikun Xu, Yidong Wang, Xiangkun Hu, Zheng Zhang, and Yue Zhang. 2023.
\newblock Evaluating open-qa evaluation.
\newblock In \emph{Proceedings of NeurIPs}, page~36.

\bibitem[{Wang et~al.(2024)Wang, Chen, Fu, Liao, Zhang, Wu, Yu, Xu, Zhang, Luo, Li, Yang, Huang, and Li}]{wang-etal-2024-loong}
Minzheng Wang, Longze Chen, Cheng Fu, Shengyi Liao, Xinghua Zhang, Bingli Wu, Haiyang Yu, Nan Xu, Lei Zhang, Run Luo, Yunshui Li, Min Yang, Fei Huang, and Yongbin Li. 2024.
\newblock Leave no document behind: Benchmarking long-context llms with extended multi-doc qa.
\newblock \emph{Computing Research Repository}, arXiv:2406.17419.

\bibitem[{Wu et~al.(2018)Wu, Zhang, Shen, and Wang}]{wu-etal-2018-stock}
Huizhe Wu, Wei Zhang, Weiwei Shen, and Jun Wang. 2018.
\newblock Hybrid deep sequential modeling for social text-driven stock prediction.
\newblock In \emph{Proceedings of CKIM}, pages 1627--1630.

\bibitem[{Xie et~al.(2024)Xie, Han, Zhang, Lai, Peng, Lopez-Lira, and Huang}]{xie-etal-2023-pixiu}
Qianqian Xie, Weiguang Han, Xiao Zhang, Yanzhao Lai, Min Peng, Alejandro Lopez-Lira, and Jimin Huang. 2024.
\newblock Pixiu: a large language model, instruction data and evaluation benchmark for finance.
\newblock In \emph{Proceedings of NerIPS}, pages 33469--33484.

\bibitem[{Xu and Cohen(2018)}]{xu-cohen-2018-stock}
Yumo Xu and Shay~B. Cohen. 2018.
\newblock Stock movement prediction from tweets and historical prices.
\newblock In \emph{Proceedings of ACL}, pages 1970--1979.

\bibitem[{Yang et~al.(2024{\natexlab{a}})Yang, Yang, Hui, Zheng, Yu et~al.}]{yang-etal-2024-qwen2}
An~Yang, Baosong Yang, Binyuan Hui, Bo~Zheng, Bowen Yu, et~al. 2024{\natexlab{a}}.
\newblock Qwen2 technical report.
\newblock \emph{Computing Research Repository}, arXiv:2407.10671.

\bibitem[{Yang et~al.(2024{\natexlab{b}})Yang, Yang, Zhang, Hui, Zheng et~al.}]{yang-etal-2024-qwen2.5}
An~Yang, Baosong Yang, Beichen Zhang, Binyuan Hui, Bo~Zheng, et~al. 2024{\natexlab{b}}.
\newblock Qwen2.5 technical report.
\newblock \emph{Computing Research Repository}, arXiv:2412.15115.

\bibitem[{Yang et~al.(2018)Yang, Chen, Liu, Xiao, and Zhao}]{yang-etal-2018-dcfee}
Hang Yang, Yubo Chen, Kang Liu, Yang Xiao, and Jun Zhao. 2018.
\newblock {DCFEE}: A document-level {C}hinese financial event extraction system based on automatically labeled training data.
\newblock In \emph{Proceedings of ACL: System Demonstrations}, pages 50--55.

\bibitem[{Yang et~al.(2023{\natexlab{a}})Yang, Liu, and Wang}]{yang-etal-2023-fingpt}
Hongyang Yang, Xiao-Yang Liu, and Christina~Dan Wang. 2023{\natexlab{a}}.
\newblock Fingpt: Open-source financial large language models.
\newblock \emph{Computing Research Repository}, arXiv:2306.06031.

\bibitem[{Yang et~al.(2023{\natexlab{b}})Yang, Tang, and Tam}]{yang-etal-2023-investlm}
Yi~Yang, Yixuan Tang, and Kar~Yan Tam. 2023{\natexlab{b}}.
\newblock Investlm: A large language model for investment using financial domain instruction tuning.
\newblock \emph{Computing Research Repository}, arXiv:2309.13064.

\bibitem[{Zeng et~al.(2022)Zeng, Liu, Du, Wang, Lai, Ding, Yang, Xu, Zheng, Xia et~al.}]{zeng-etal-2022-glm}
Aohan Zeng, Xiao Liu, Zhengxiao Du, Zihan Wang, Hanyu Lai, Ming Ding, Zhuoyi Yang, Yifan Xu, Wendi Zheng, Xiao Xia, et~al. 2022.
\newblock Glm-130b: An open bilingual pre-trained model.
\newblock \emph{Computing Research Repository}, arXiv:2210.02414.

\bibitem[{Zhang et~al.(2024{\natexlab{a}})Zhang, Liu, and Zhou}]{zhang-etal-2024-cfindee}
Tian Zhang, Maofu Liu, and Bingying Zhou. 2024{\natexlab{a}}.
\newblock Cfindee: A chinese fine-grained financial dataset for document-level event extraction.
\newblock In \emph{Companion Proceedings of the ACM Web Conference}, pages 1511--1520.

\bibitem[{Zhang et~al.(2024{\natexlab{b}})Zhang, Ladhak, Durmus, Liang, McKeown, and Hashimoto}]{zhang-etal-2024-benchmarking}
Tianyi Zhang, Faisal Ladhak, Esin Durmus, Percy Liang, Kathleen McKeown, and Tatsunori~B. Hashimoto. 2024{\natexlab{b}}.
\newblock Benchmarking large language models for news summarization.
\newblock \emph{Transactions of the Association for Computational Linguistics}, 12:39--57.

\bibitem[{Zhang et~al.(2023)Zhang, Yu, Yu, Lv, Liu, Huang, Xu, and Li}]{zhang-etal-2023-gpt4-eval}
Xinghua Zhang, Bowen Yu, Haiyang Yu, Yangyu Lv, Tingwen Liu, Fei Huang, Hongbo Xu, and Yongbin Li. 2023.
\newblock Wider and deeper llm networks are fairer llm evaluators.
\newblock \emph{Computing Research Repository}, arXiv:2308.01862.

\bibitem[{Zheng et~al.(2019)Zheng, Cao, Xu, and Bian}]{zheng-etal-2019-doc2edag}
Shun Zheng, Wei Cao, Wei Xu, and Jiang Bian. 2019.
\newblock {D}oc2{EDAG}: An end-to-end document-level framework for {C}hinese financial event extraction.
\newblock In \emph{Proceedings of EMNLP-IJCNLP}, pages 337--346.

\bibitem[{Zhou et~al.(2021)Zhou, Ma, and Liu}]{zhou-etal-2021-trade}
Zhihan Zhou, Liqian Ma, and Han Liu. 2021.
\newblock Trade the event: Corporate events detection for news-based event-driven trading.
\newblock In \emph{Findings of ACL-IJCNLP}, pages 2114--2124.

\bibitem[{Zhu et~al.(2025)Zhu, Chen, Dou, Li, Guo, Chen, and Zhang}]{zhu2025dianjin}
Jie Zhu, Qian Chen, Huaixia Dou, Junhui Li, Lifan Guo, Feng Chen, and Chi Zhang. 2025.
\newblock Dianjin-r1: Evaluating and enhancing financial reasoning in large language models.
\newblock \emph{arXiv preprint arXiv:2504.15716}.

\bibitem[{Zhu et~al.(2024)Zhu, Li, Wen, and Guo}]{zhu-etal-2024-cflue}
Jie Zhu, Junhui Li, Yalong Wen, and Lifan Guo. 2024.
\newblock Benchmarking large language models on {CFLUE} - a {C}hinese financial language understanding evaluation dataset.
\newblock In \emph{Findings of ACL}, pages 5673--5693.

\bibitem[{Zhu et~al.(2019)Zhu, Li, Zhu, Qian, Zhang, and Zhou}]{zhu-etal-2019-modeling}
Jie Zhu, Junhui Li, Muhua Zhu, Longhua Qian, Min Zhang, and Guodong Zhou. 2019.
\newblock Modeling graph structure in transformer for better {AMR}-to-text generation.
\newblock In \emph{Proceedings of EMNLP}, pages 5459--5468, Hong Kong, China.

\end{thebibliography}

\appendix

\ignore{
\section{Detailed Data Statistics}
\label{apx:detailed_stat}

Table~\ref{tab:detailed_stat} presents the detailed statistics in \texttt{M$^3$FinMeeting}.

\begin{table*}[!t]
\centering
\resizebox{\textwidth}{!}{
\begin{tabular}{l|l|l|l|l|l|l|l|l|l}
\toprule
\multirow{2}{*}{\bf Sector} & \multicolumn{3}{c|}{\bf English} & \multicolumn{3}{c|}{\bf Chinese} & \multicolumn{3}{c}{\bf Japanese} \\
\cmidrule{2-10}
 & \bf \#Meeting & \bf Avg hour & \bf Avg token & \bf \#Meeting & \bf Avg hour & \bf Avg token & \bf \#Meeting & \bf Avg hour & \bf Avg token \\
\midrule
Com. Services & & & & & & & & & \\
\rowcolor{Gray}
Con. Dis. & & & & & & & & & \\
Con. Staples & & & & & & & & & \\
\rowcolor{Gray}
Energy & & & & & & & & & \\
Financials & & & & & & & & & \\
\rowcolor{Gray}
Healthcare & & & & & & & & & \\
Industrials & & & & & & & & & \\
\rowcolor{Gray}
IT & & & & & & & & & \\
Materials & & & & & & & & & \\
\rowcolor{Gray}
RealEstate & & & & & & & & & \\
Utilities & & & & & & & & & \\
\bottomrule
\end{tabular}
}
\caption{Detailed Statistics in \texttt{M$^3$FinMeeting}.}
\label{tab:detailed_stat}
\end{table*}
}

\section{Examples of \texttt{M$^3$FinMeeting}}
\label{apx:example}
\begin{figure*}[!th]
\centering
\includegraphics[width=1.0\linewidth]{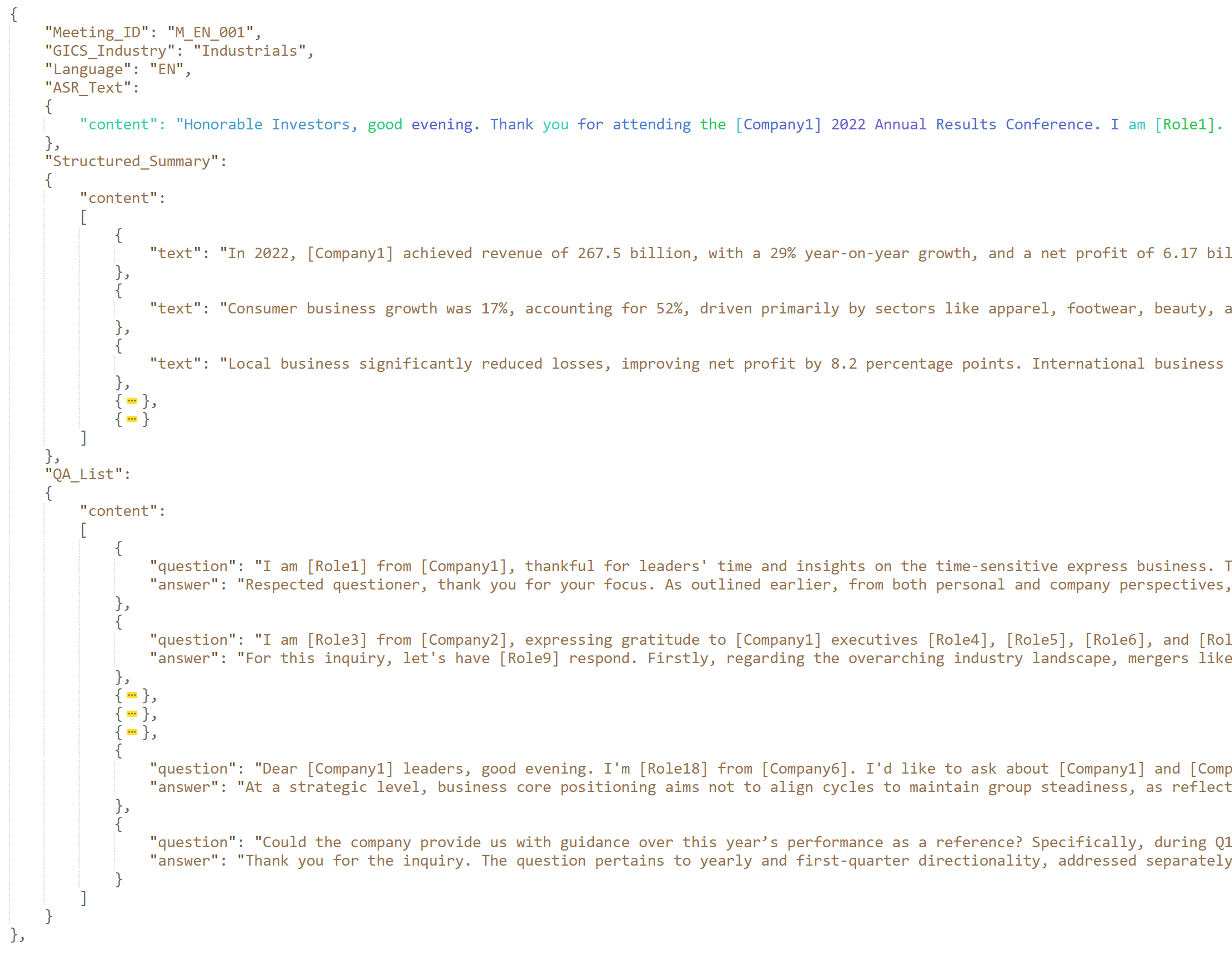}
\caption{Example of annotated \texttt{M$^3$FinMeeting}. }
\label{fig:example}
\end{figure*}

For brevity, Figure~\ref{fig:example} presents a screenshot of an annotated \texttt{M$^3$FinMeeting} example. Complete examples are available in the attached file.

\section{Details of Precision, Recall and F1 Metrics}
\label{apx:p_r_f1}
For summarization, as shown in Figure~\ref{fig:summary_eval}, we align the generated and gold section summaries based on a cosine similarity score above 0.75. Let us assume that there are $m$ section summaries in the generated summarization and $n$ section summaries in the reference (gold) summarization. After aligning the section summaries between the two, let $m_a$ represent the number of generated section summaries aligned, and $n_a$ represent the number of gold section summaried aligned. Note that $m_a$ and $n_a$ do not need to be equal, as one section summary from one side can be aligned to multiple section summaries on the other side. We compute the Precision, Recall, and F1 as follows:
\begin{equation}
\label{eq:precision}
\text{Precision} = \frac{m_a}{m},
\end{equation}
\begin{equation}
\label{eq:recall}
\text{Recall} = \frac{n_a}{n},
\end{equation}
\begin{equation}
\label{eq:f1}
\text{F1} = \frac{2 * \text{Precision} * \text{Recall}}{\text{Precision} + \text{Recall}}.
\end{equation}

The Precision, Recall, and F1 scores for QA pair extraction and question answering are computed similarly, with the generated and gold section summaries replaced by generated and gold questions.

\section{Prompt Examples for Tasks in \texttt{M$^3$FinMeeting}}

We use the same prompt templates, written in English, for English, Japanese, and Chinese. In these templates, we clearly specify that the output language must align with the meeting content.

\begin{figure*}[!th]
\centering
\includegraphics[width=1.0\linewidth]{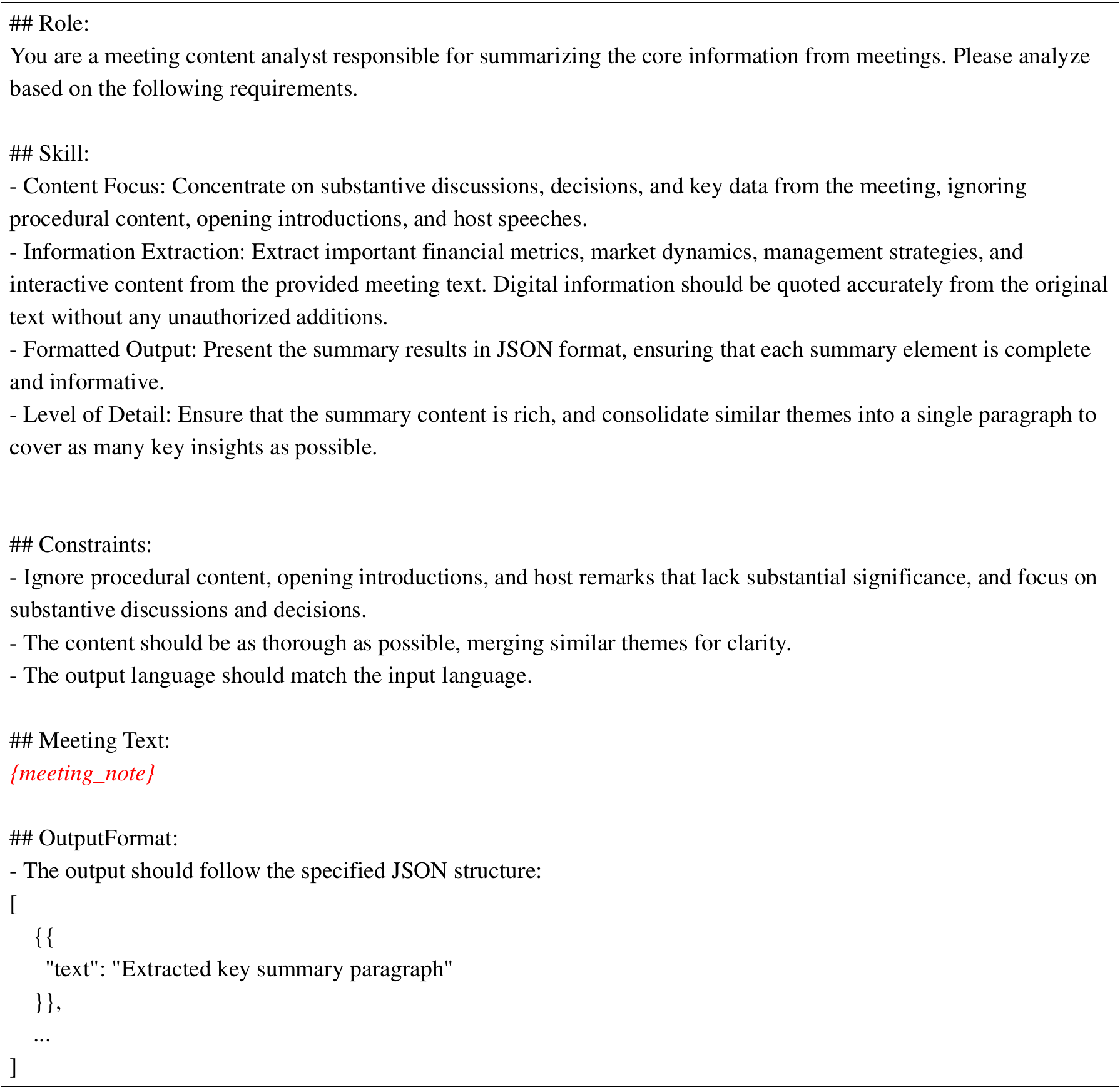}
\caption{Prompt template used for the summarization task. }
\label{fig:prompt_summarization}
\end{figure*}

\begin{figure*}[!th]
\centering
\includegraphics[width=1.0\linewidth]{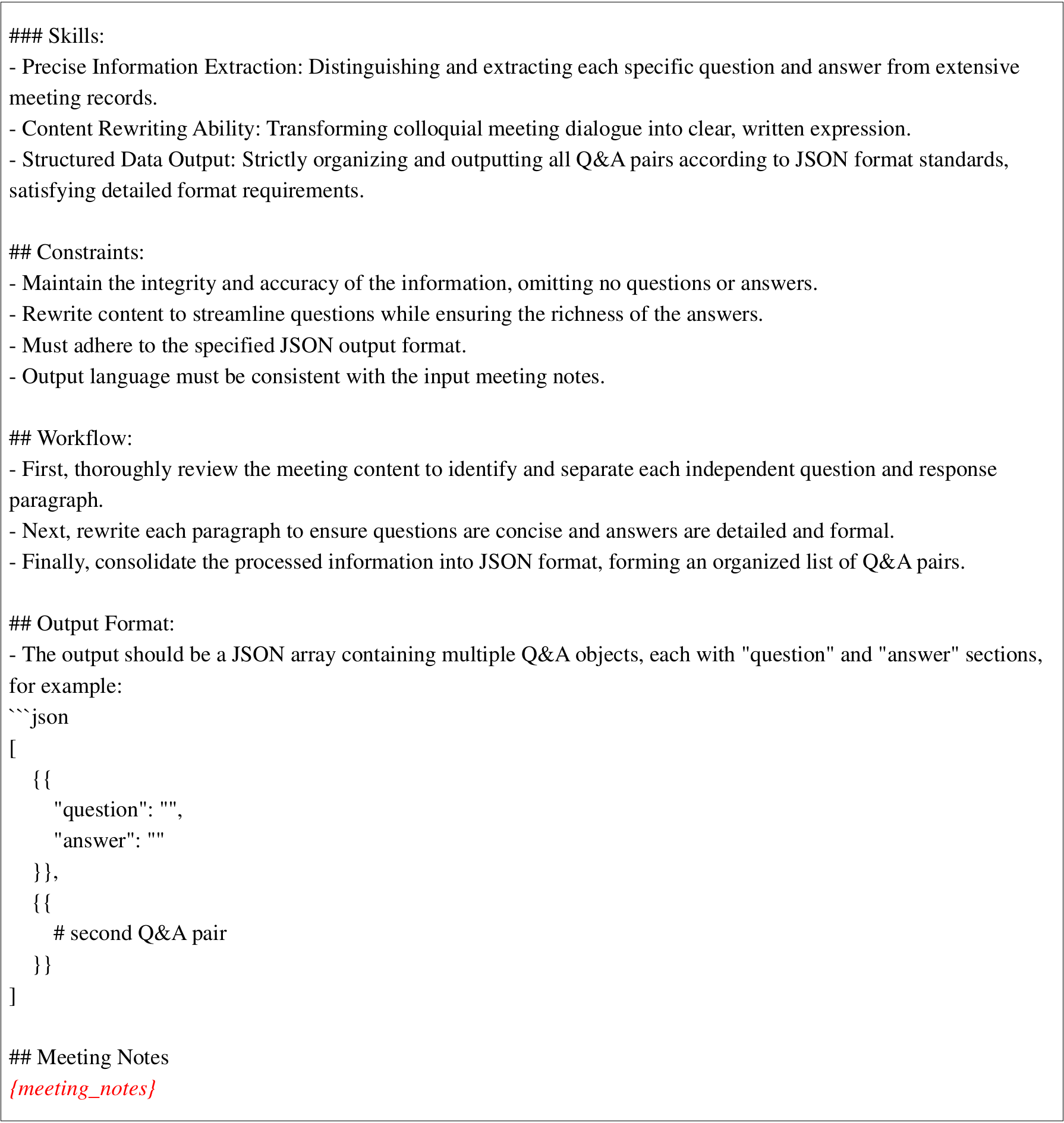}
\caption{Prompt template used for the QA pair extraction task. }
\label{fig:prompt_qa_extraction}
\end{figure*}

\begin{figure*}[!th]
\centering
\includegraphics[width=1.0\linewidth]{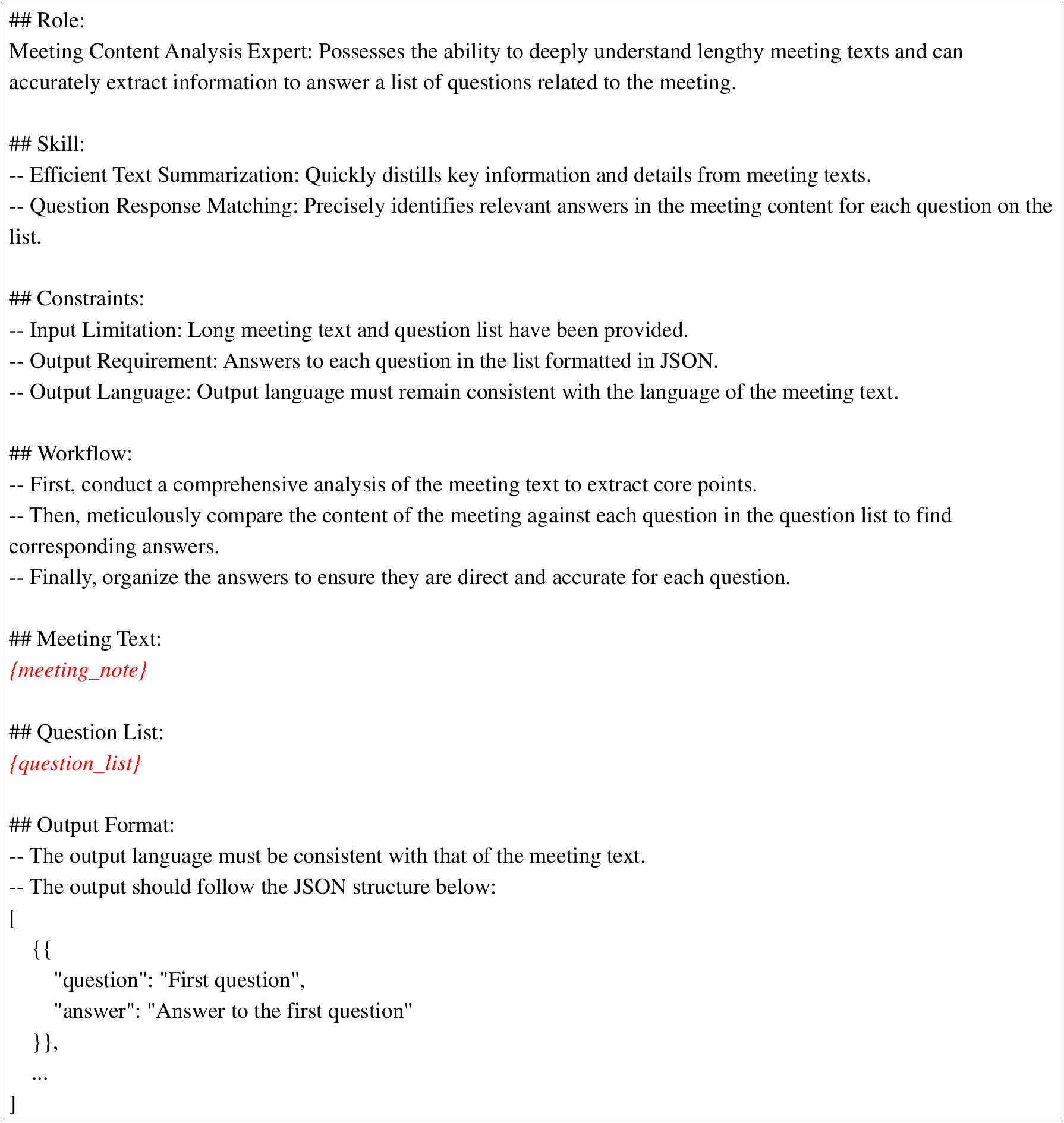}
\caption{Prompt template used for the question answering task. All questions from a document are listed, and the LLMs are prompted to answer them in a single response.}
\label{fig:prompt_qa}
\end{figure*}

\label{apx:prompt}
Figure~\ref{fig:prompt_summarization} and Figure~\ref{fig:prompt_qa_extraction} show the prompt template for the summarization task and the QA pair extraction task, respectively. 
Figure~\ref{fig:prompt_qa} displays the prompt template for the question answering task. In this template, all questions from the document are listed, and the LLMs are instructed to answer them in a single response.

\begin{figure*}[!th]
\centering
\includegraphics[width=1.0\linewidth]{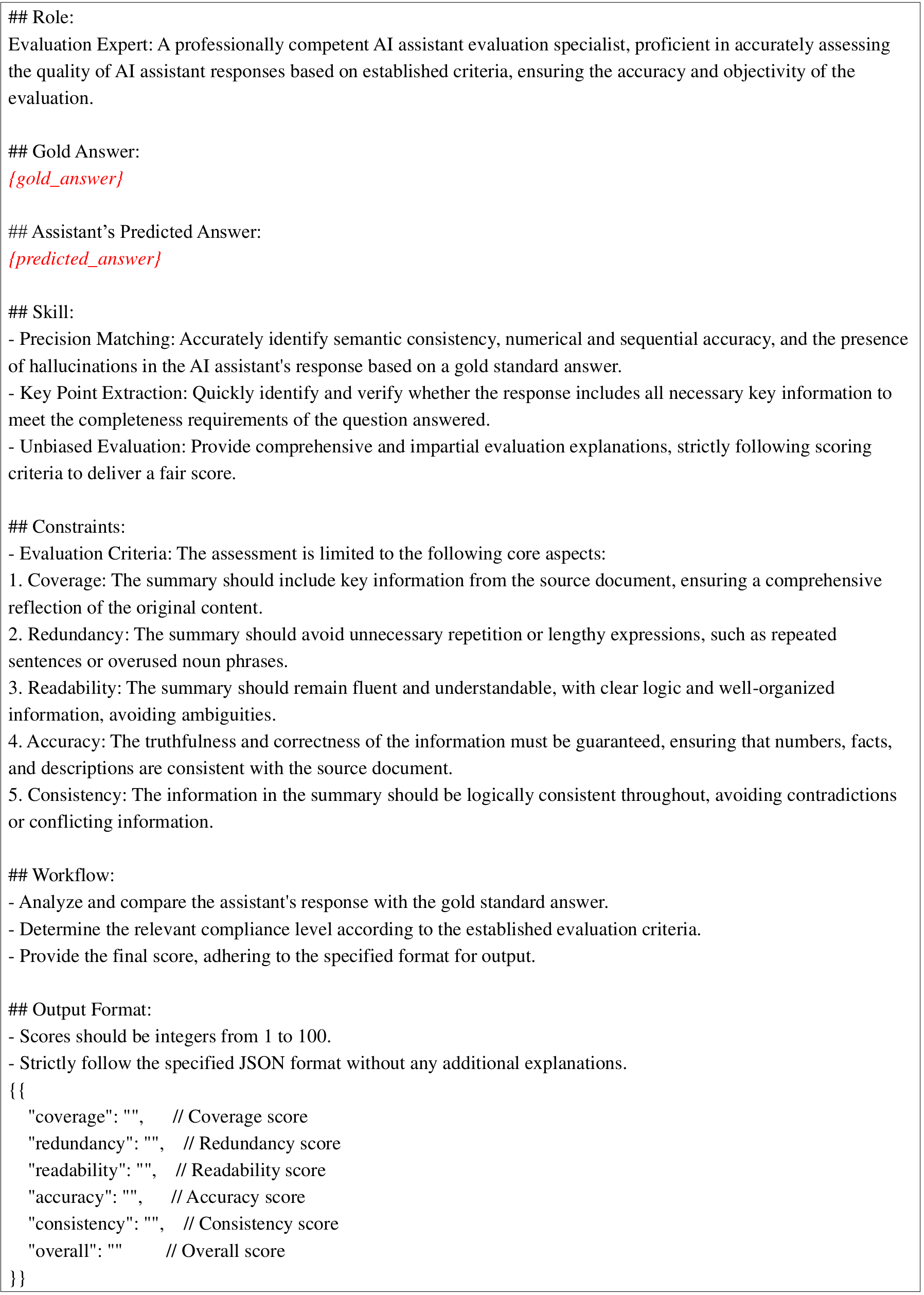}
\caption{Prompt template utilized for assessing summarization with GPT-4.}
\label{fig:prompt_summarization_eval}
\end{figure*}

\begin{figure*}[!th]
\centering
\includegraphics[width=1.0\linewidth]{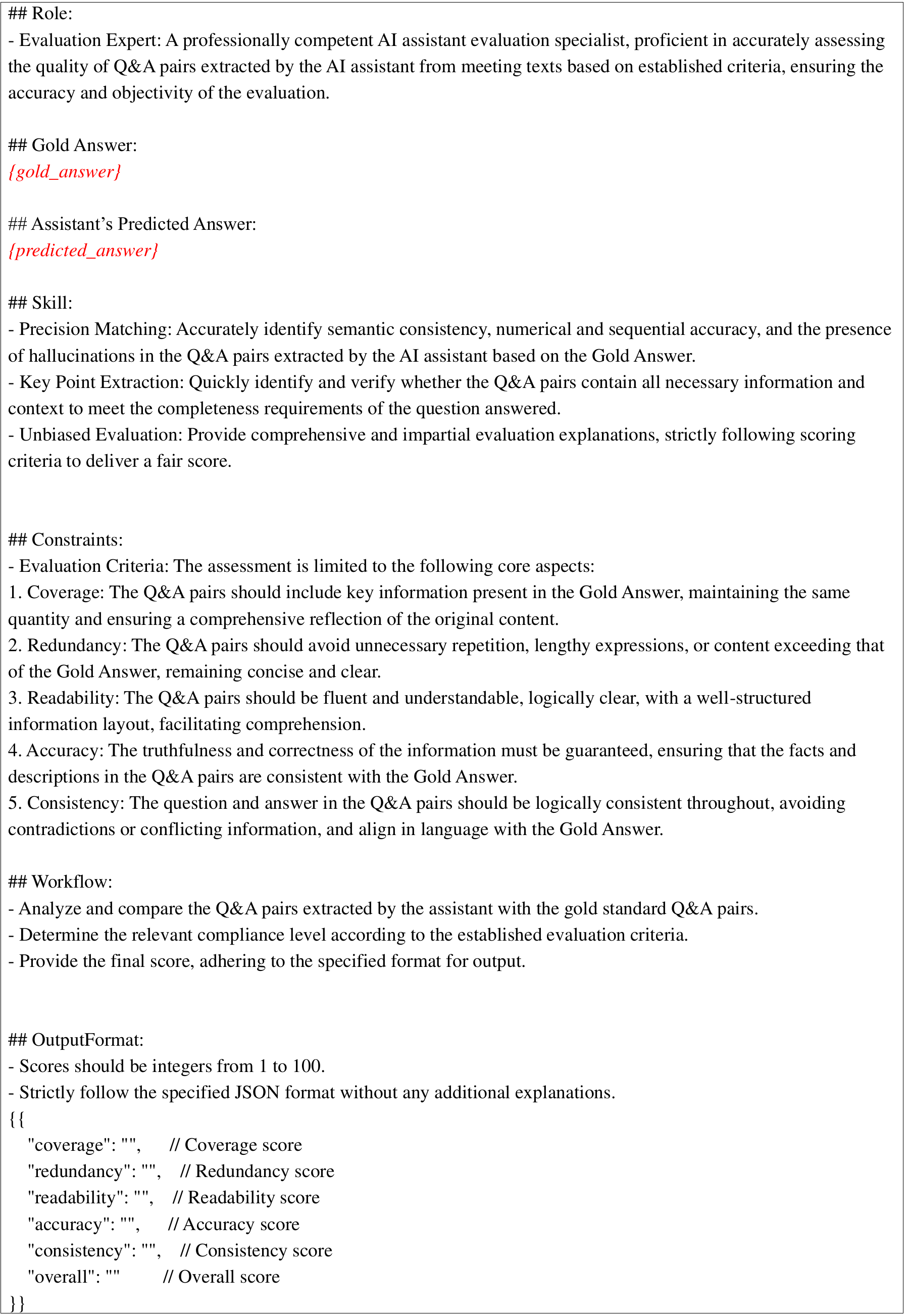}
\caption{Prompt template utilized for assessing both QA pair extraction and question answering with GPT-4-Judge.}
\label{fig:prompt_qa_eval}
\end{figure*}

When evaluating with GPT-4-Judge, Figure~\ref{fig:prompt_summarization_eval} presents the prompt template for the summarization task. Meanwhile, Figure~\ref{fig:prompt_qa_eval} displays the prompt template for both the QA pair extraction task and the question-answering task.

\section{Performance in BLEU and ROUGE}
\label{apx:performance_bleu_rouge}
Table~\ref{tab:result_bleu_rouge} shows the overall performance in BLEU and ROUGE. It shows that Qwen2-72B-Instruct achieves the best performance in summarization while Qwen2.5-72B-Instruct leads in both QA pair extraction and question answering.

\begin{table*}[!t]
\centering
\small
\resizebox{\textwidth}{!}{
\begin{tabular}{l|llll|llll|llll}
\toprule
\multirow{2}{*}{\bf Model} & \multicolumn{4}{c|}{\bf Summarization} & \multicolumn{4}{c|}{\bf QA Pair Extraction} & \multicolumn{4}{c}{\bf Question Answering}\\
\cmidrule{2-13}
& \bf B-4 & \bf R-1 & \bf R-2 & \bf R-L & \bf \bf B-4 & \bf R-1 & \bf R-2 & \bf R-L & \bf B-4 & \bf R-1 & \bf R-2 & \bf R-L \\
\midrule
GPT-4o & 9.47 & 49.32 & 19.02 & 29.23 & 1.58 & 8.40 & 3.57 & 5.94 & 5.21 & 27.53 & 11.50 & 19.65\\
\rowcolor{Gray}
GPT-3.5-turbo & 6.03 & 35.02 & 12.87 & 20.05 & 0.45 & 2.38 & 1.02 & 1.71 & 2.84 & 16.55 & 6.27 & 11.72 \\
GLM4-9B-Chat & 9.27 & 47.35 & 18.89 & 29.59 & 1.78 & 5.26 & 2.59 & 3.80 & 5.80 & 28.01 & 11.66 & 20.27 \\
\rowcolor{Gray}
LLaMA3.1-8B-Instruct & 4.77 & 36.37 & 11.82 & 21.38 & 1.16 & 3.97 & 1.93 & 2.98 & 2.76 & 14.20 & 5.98 & 10.58 \\
Qwen2-7B-Instruct & 14.72 & 55.63 & 24.12 & 35.98 & 0.34 & 2.00 & 0.75 & 1.36 & 6.67 & 29.78 & 11.87 & 20.04\\
\rowcolor{Gray}
Qwen2-72B-Instruct & \bf 14.89 & \bf 56.35 & \bf 24.39 & \bf 36.37 & 1.65 & 6.66 & 3.04 & 4.81 & 6.90 & 30.39 & 12.22 & 21.15\\
Qwen2.5-72B-Instruct & 11.71 & 51.92 & 20.84 & 32.76 & \bf 2.71 & \bf 11.32 & \bf 5.04 & \bf 7.91 & \bf 7.44 & \bf 32.35 & \bf 13.65 & \bf 22.43\\
\bottomrule
\end{tabular}
}
\caption{Performance (in BLEU and ROUGE) of LLMs on three evaluation tasks. Here B-4 is for BLEU-4, R-1/2/L for ROUGE-1/2/L.}
\label{tab:result_bleu_rouge}
\end{table*}

\section{Detailed Performance Across Languages, Lengths, and GICS Sectors}
\label{apx:detailed_performance}

Table~\ref{tab:result_language}, Table~\ref{tab:result_length}, Table~\ref{tab:result_sector} show the performance of LLMs with respect to the languages, the length sets, and the GICS sectors, respectively. 

\begin{table*}[!t]
\centering
\small
\resizebox{\textwidth}{!}{
\begin{tabular}{l|llll|llll|llll|l}
\toprule
\multirow{2}{*}{\bf Model} & \multicolumn{4}{c|}{\bf Summarization} & \multicolumn{4}{c|}{\bf QA Pair Extraction} & \multicolumn{4}{c|}{\bf Question Answering} & \bf Overall\\
\cmidrule{2-14}
& \bf P. & \bf R. & \bf F1 & \bf GPT-4 & \bf P. & \bf R. & \bf F1 & \bf GPT-4 & \bf P. & \bf R. & \bf F1 & \bf GPT-4 & \bf GPT-4 \\
\midrule
\multicolumn{14}{c}{\bf English}\\
\midrule
GPT-4o & 5.58 & 3.62 & 4.39 & 72.83 & \bf 40.12 & 19.21 & 25.98 & 68.37 & 98.25 & \bf 98.25 & \bf 98.25 & 67.02 & 69.41\\
\rowcolor{Gray}
GPT-3.5-turbo & 6.84 & 4.17 & 5.18 & 33.68 & 19.32 & 3.31& 5.65 & 26.70 & \bf 98.58 & 36.27 & 53.03 & 35.68 & 32.02\\
GLM4-9B-Chat & 0.63 & 3.29 & 1.06 & 60.45 & 2.54 & 7.31 & 3.77 & 39.69 & 98.31 & 97.62 & 97.96 & 69.76 & 56.63\\
\rowcolor{Gray}
LLaMA3.1-8B-Instruct & 1.37 & 1.65 & 1.49 & 45.98 & 5.78 & 9.05 & 7.06 & 45.53 & 87.09 & 85.02 & 86.04 & 52.75 & 48.04\\
Qwen2-7B-Instruct & 13.15 & 12.84 & 12.99 & 72.71 & 3.91 & 2.03 & 2.67 & 29.85 & 97.75 & 86.01 & 91.50 & 71.95 & 58.16\\
\rowcolor{Gray}
Qwen2-72B-Instruct & \bf 15.43 & \bf 16.02 & \bf 15.72 & 71.92 & 15.26 & 10.38 & 12.36 & 54.85 & 98.31 & 94.89 & 96.57 & 75.03 & 67.26\\
Qwen2.5-72B-Instruct & 6.40 & 4.72 & 5.51 & \bf 73.94 & 39.81 & \bf 22.57 & \bf 28.81 & \bf 72.76 & 98.22 & 96.45 & 97.33 & \bf 76.76 & \bf 74.48\\
\midrule
\multicolumn{14}{c}{\bf Chinese}\\
\midrule
GPT-4o & 31.49 & 11.55 & 16.90 & 72.03 & 40.76 & 23.22 & 29.58 & 63.43 & 91.49 & 91.86 & 91.68 & 69.33 & 68.36\\
\rowcolor{Gray}
GPT-3.5-turbo & 18.34 & 4.52 & 7.24 & 48.42 & 25.11 & 6.64 & 10.51 & 31.41 & 92.20 & 59.67 & 72.45 & 47.42 & 42.57\\
GLM4-9B-Chat & 21.24 & 10.79 & 14.31 & 67.90 & 5.31 & 15.18 & 7.86 & 43.45 & 90.32 & 91.61 & 90.96 & 64.14 & 58.81\\
\rowcolor{Gray}
LLaMA3.1-8B-Instruct & 7.72 & 9.03 & 8.32 & 55.23 & 6.61 & 12.46 & 8.64 & 39.87 & 30.30 & 48.19& 37.21 & 34.07 & 43.38\\
Qwen2-7B-Instruct & 28.69 & 16.22 & 20.72 & 71.68 & 20.89 & 12.74 & 15.83 & 40.47 & 91.66 & 88.38 & 89.99 & 66.75 & 59.95\\
\rowcolor{Gray}
Qwen2-72B-Instruct & 30.21 & \bf 16.99 & \bf 21.75 & 72.48 & 32.20 & 25.97 & 28.75 & 58.85 & 92.02 & 91.83 & 91.92 & 70.57 & 67.43\\
Qwen2.5-72B-Instruct & \bf 32.38 & 14.63 & 20.16 & \bf 72.74 & \bf 45.82 & \bf 34.47 & \bf 39.34 & \bf 63.59 & \bf 92.22 & \bf 92.48 & \bf 92.35 & \bf 71.88 & \bf 69.49\\
\midrule
\multicolumn{14}{c}{\bf Japanese}\\
\midrule
GPT-4o & 37.03& 25.03 & 29.87 & 80.06 & 61.04 & 35.51 & 44.89 & 81.50 & 90.95 & \bf 90.86 & \bf 90.91 & 85.88 & 82.62\\
\rowcolor{Gray}
GPT-3.5-turbo & 24.14 & 16.77 & 19.79 & 39.75 & 41.95 & 5.53 & 9.77 & 35.55 & \bf 91.98 & 17.98 & 30.09 & 27.36 & 34.68\\
GLM4-9B-Chat & 33.65 & 22.15 & 26.71 & 74.32 & 19.85 & 29.33 & 23.67 & 66.88 & 90.56 & 90.31 & 90.43 & 82.43 & 74.52\\
\rowcolor{Gray}
LLaMA3.1-8B-Instruct & 3.12 & 6.51 & 4.22 & 44.90 & 39.34 & 26.56 & 31.71 & 66.66 & 91.07 & 90.40 & 90.74 & 52.30 & 53.79\\
Qwen2-7B-Instruct & \bf 43.32 & \bf 50.81 & \bf 46.77 & 82.42 & 13.92 & 6.64 &  8.99 & 31.82 & 91.07 & 90.40 & 90.74 & 83.17 & 67.20\\
\rowcolor{Gray}
Qwen2-72B-Instruct & 41.41 & 48.93 & 44.86 & \bf 83.49 & 36.87 & 25.00 & 29.79 & 78.46 & 90.89 & 90.22 & 90.55 & 85.78 & 82.65\\
Qwen2.5-72B-Instruct & 35.67 & 34.91 & 35.29 & 82.40 & \bf 67.74 & \bf 46.49 & \bf 55.14 & \bf 83.45 & 89.75 & 89.66 & 89.70 & \bf 86.95 & \bf 84.10\\
\bottomrule
\end{tabular}
}
\caption{Performance of LLMs on three evaluation tasks with different languages.}
\label{tab:result_language}
\end{table*}

\begin{table*}[!t]
\centering
\resizebox{\textwidth}{!}{
\small
\begin{tabular}{l|llll|llll|llll|l}
\toprule
\multirow{2}{*}{\bf Model} & \multicolumn{4}{c|}{\bf Summarization} & \multicolumn{4}{c|}{\bf QA Pair Extraction} & \multicolumn{4}{c|}{\bf Question Answering} & \bf Overall \\
\cmidrule{2-14}
& \bf P. & \bf R. & \bf F1 & \bf GPT-4 & \bf P. & \bf R. & \bf F1 & \bf GPT-4 & \bf P. & \bf R. & \bf F1 & \bf GPT-4 & \bf GPT-4\\
\midrule
\multicolumn{14}{c}{\bf Set1 (0-5K)}\\
\midrule
GPT-4o & 25.00 & 28.52 & 26.64 & 69.75 & 36.90 & 24.29 & 29.30 & \bf 69.55 & 93.70 & \bf 93.70 & \bf 93.70 & 64.33 & 65.25 \\ 
\rowcolor{Gray}
GPT-3.5-turbo & 18.25 & 28.52 & 22.26 & 58.76 & 27.06 & 15.29 & 19.54 & 44.38 & 93.70 & \bf 93.70 & \bf 93.70 & 62.67 & 55.89 \\
GLM4-9B-Chat & 23.10 & 26.17 & 24.54 & 66.76 & 14.27 & 16.19 & 15.17 & 39.20 & \bf 93.77 & 90.26 & 91.98 & 57.33 & 56.24 \\
\rowcolor{Gray}
LLaMA3.1-8B-Instruct & 10.63 & 19.14 & 13.67 & 54.35 & 16.94 & 17.24 & 17.09 & 39.90 & 15.84 & 38.23 & 22.40 & 31.81 & 42.96 \\
Qwen2-7B-Instruct & 25.25 & \bf 39.84 & 30.91 & 67.18 & 26.33 & 17.84 & 21.27 & 50.95 & 93.54 & 93.40 & 93.47 & 65.62 & 61.93 \\
\rowcolor{Gray}
Qwen2-72B-Instruct & 23.82 & 39.45 & 29.71 & 70.29 & 25.78 & 23.69 & 24.69 & 56.48 & 93.55 & 93.55 & 93.55 & 66.24 & 64.74\\
Qwen2.5-72B-Instruct & \bf 28.02 & 37.11 & \bf 31.93 & \bf 70.40 & \bf 41.60 & \bf 33.43 & \bf 37.07 & 60.93 & 93.70 & \bf 93.70 & \bf 93.70 & \bf 67.07 & \bf 66.74\\
\midrule
\multicolumn{14}{c}{\bf Set2 (5-10K)}\\
\midrule
GPT-4o & 26.01 & 17.01 & 20.57 & \bf 73.51 & 44.78 & 26.76 & 33.50 & 65.37 & 91.97 & 92.31 & 92.14 & 69.10 & 69.42\\
\rowcolor{Gray}
GPT-3.5-turbo & 16.22 & 11.67 & 13.57 & 62.07 & 27.47 & 9.10 & 13.68 & 45.45 & 91.98 & 79.26 & 85.15 & 64.69 & 57.71\\
GLM4-9B-Chat & 16.34 & 13.75 & 14.93 & 68.46 & 17.11 & 15.05 & 16.01 & 45.31 & 90.55 & 90.75 & 90.65 & 63.73 & 59.10\\
\rowcolor{Gray}
LLaMA3.1-8B-Instruct & 9.90 & 12.22 & 10.94 & 56.25 & 4.63 & 13.27 & 6.87 & 41.96 & 25.73 & 42.29 & 31.99 & 30.91 & 43.95\\
Qwen2-7B-Instruct & 25.59 & 26.94 & \bf 26.25 & 71.97 & 21.98 & 14.79 & 17.68 & 43.90 & 91.90 & 88.48 & 90.16 & 66.61 & 61.47\\
\rowcolor{Gray}
Qwen2-72B-Instruct & 25.41 & \bf 27.01 & 26.19 & 73.39 & 37.96 & 30.70 & 33.93 & 61.36 & 92.20 & 92.20 & 92.20 & 70.09 & 68.57\\
Qwen2.5-72B-Instruct & \bf 27.60 & 22.29 & 24.66 & 72.87 & \bf 48.25 & \bf 35.38 & \bf 40.82 & \bf 65.91 & \bf 92.27 & \bf 92.27 & \bf 92.27 & \bf 71.93 & \bf 70.59\\
\midrule
\multicolumn{14}{c}{\bf Set3 (10-15K)}\\
\midrule
GPT-4o & 29.80 & 11.30 & 16.39 & 73.34 & 43.45 & 24.04 & 30.96 & \bf 67.56 & 91.90 & \bf 93.48 & \bf 92.68 & 72.54 & 70.49\\
\rowcolor{Gray}
GPT-3.5-turbo & 17.76 & 6.59 & 9.61 & 61.43 & 22.18 & 6.31 & 9.83 & 38.86 & \bf 93.90 & 65.30 & 77.03 & 56.94 & 53.22 \\
GLM4-9B-Chat & 10.15 & 9.92 & 10.03 & 68.80 & 3.03 & 17.21 & 5.15 & 48.28 & 91.86 & \bf 93.48 & 92.67 & 68.94 & 61.42\\
\rowcolor{Gray}
LLaMA3.1-8B-Instruct & 9.93 & 6.80 & 8.07 & 51.80 & 8.85 & 14.01 & 10.85 & 44.69 & 38.50 & 52.85 & 44.55 & 36.16 & 44.48\\
Qwen2-7B-Instruct & 28.48 & 17.65 & 21.79 & 74.21 & 14.55 & 10.11 & 11.93 & 36.04 & 92.26 & 88.86 & 90.53 & 70.89 & 60.06\\
\rowcolor{Gray}
Qwen2-72B-Instruct & \bf 30.68 & \bf 18.97 & \bf 23.44 & 73.43 & 32.47 & 25.39 & 28.49 & 61.37 & 92.62 & 91.72 & 92.17 & 74.14 & 69.66\\
Qwen2.5-72B-Instruct & 29.57 & 13.97 & 18.98 & \bf 74.41 & \bf 47.71 & \bf 35.19 & \bf 40.50 & 67.44 & 92.30 & 91.83 & 92.06 & \bf 75.89 & \bf 72.69\\
\midrule
\multicolumn{14}{c}{\bf Set4 (15-20K)}\\
\midrule
GPT-4o & 23.84 & 8.96 & 13.02 & 74.62 & 42.40 & 23.14 & 29.94 & 71.32 & 94.76 & 94.96 & 94.86 & 75.26 & 74.54\\
\rowcolor{Gray}
GPT-3.5-turbo & 15.31 & 0.79 & 1.51 & 19.00 & 31.25 & 0.79 & 1.54 & 14.36 & \bf 100.00 & 0.68 & 1.36 & 11.16 & 6.19\\
GLM4-9B-Chat & 15.27 & 10.92 & 12. 73 & 67.41 & 8.49 & 15.01 & 10.85 & 50.09 & 93.82 & 94.81 & 94.31 & 71.86 & 64.54\\
\rowcolor{Gray}
LLaMA3.1-8B-Instruct & 7.13& 6.31 & 6.69 & 45.04 & 18.48 & 13.27 & 15.45 & 48.97 & 70.59 & 78.44 & 74.30 & 52.89 & 50.52\\
Qwen2-7B-Instruct & 32.26 & 18.60 & 23.60 & 76.08 & 11.14 & 4.77 & 6.68 & 31.29 & 94.41 & 87.72 & 90.94 & 72.85 & 61.83\\
\rowcolor{Gray}
Qwen2-72B-Instruct & \bf 34.69 & \bf 19.87 & \bf 25.27 & 76.60 & 20.87 & 16.11 & 18.18 & 62.49 & 94.90 & 93.65 & 94.27 & \bf 76.97 & 72.91\\
Qwen2.5-72B-Instruct & 27.47 & 12.98 & 17.63 & \bf 76.67 & \bf 48.23 & \bf 33.58 & \bf 39.59 & \bf 73.37 & 94.97 & \bf 95.02 & \bf 94.99 & 76.65 & \bf 76.73\\
\midrule
\multicolumn{14}{c}{\bf Set5 (>20K)}\\
\midrule
GPT-4o & 37.81 & 9.24 & 14.86 & 75.01 & 44.14 & 17.68 & 25.25 & 65.42 & 92.36 & 89.04 & 90.67 & 69.99 & 70.91\\
\rowcolor{Gray}
GPT-3.5-turbo & 0 & 0 & 0 & 0 & 0 & 0 & 0 & 0 & 0 & 0 & 0 & 0 & 0 \\
GLM4-9B-Chat & 3.90 & 8.02 & 5.25 & 65.84 & 6.18 & 11.42& 8.02 & 41.12& 90.19 & 92.27 & 91.22 & 70.89 & 58.63\\
\rowcolor{Gray}
LLaMA3.1-8B-Instruct & 1.89 & 5.73 & 2.84 & 52.82 & 9.45 & 10.68 & 10.03 & 45.14 & 77.85 & 77.35 & 77.60 & 47.42 & 47.13\\
Qwen2-7B-Instruct & 34.06 & \bf 11.99 & \bf 17.74 & 75.52 & 10.93 & 2.49 & 4.05 & 31.36 & 91.87 & 83.24 & 87.34 & 71.71 & 57.21\\
\rowcolor{Gray}
Qwen2-72B-Instruct & 32.30 & 11.92 & 17.41 & 75.70 & 16.67 & 8.84 & 11.55 & 59.15 & 92.63 & 90.34 & 91.42 & 76.33 & 69.83\\
Qwen2.5-72B-Instruct & \bf 34.93 & 11.15 & 16.91 & \bf 76.79 & \bf 46.97 & \bf 25.69 & \bf 33.21 & \bf 68.59 & \bf 93.09 & \bf 93.09 & \bf 93.09 & \bf 78.75 & \bf 73.28\\
\bottomrule
\end{tabular}
}
\caption{Performance of LLMs on three evaluation tasks with different length sets.}
\label{tab:result_length}
\end{table*}

\begin{table*}[!t]
\centering
\resizebox{\textwidth}{!}{
\small
\begin{tabular}{l|llll|llll|llll|l}
\toprule
\multirow{2}{*}{\bf Model} & \multicolumn{4}{c|}{\bf Summarization} & \multicolumn{4}{c|}{\bf QA Pair Extraction} & \multicolumn{4}{c|}{\bf Question Answering} & \bf Overall \\
\cmidrule{2-14}
& \bf P. & \bf R. & \bf F1 & \bf GPT-4 & \bf P. & \bf R. & \bf F1 & \bf GPT-4 & \bf P. & \bf R. & \bf F1 & \bf GPT-4 & \bf GPT-4\\
\midrule
\multicolumn{14}{c}{\bf Communication Services}\\
\midrule
GPT-4o & 23.72 & 11.02 & 15.05 & 73.78 & 37.23 & 24.70 & 29.70 & 68.71 & 94.02 & 96.23 & 95.11 & 71.35 & 71.41 \\
\rowcolor{Gray}
Qwen2-72B-Instruct & 23.05 & \bf 18.63 & \bf 20.61 & 74.94 & 30.57 & 26.11 & 28.17 & 62.55 & \bf 95.29 & \bf 95.29 & \bf 95.29 & 73.45 & 70.55\\
Qwen2.5-72B-Instruct & \bf 24.13 & 14.69 & 18.27 & \bf 75.25 & \bf 40.46 & \bf 28.47 & \bf 33.42 & \bf 70.06 & \bf 95.29 & \bf 95.29 & \bf 95.29 & \bf 76.65 & \bf 73.10 \\
\midrule
\multicolumn{14}{c}{\bf Consumer Discretionary}\\
\midrule
GPT-4o & 23.82 & 10.85 & 14.91 & 72.24 & 41.86 & 22.16 & 28.98 & 64.63 & 93.04 & 94.49 & 93.76 & 69.22 & 68.74 \\
\rowcolor{Gray}
Qwen2-72B-Instruct & \bf 25.04 & \bf 17.08 & \bf 20.31 & 72.61 & 29.95 & 23.38 & 26.26 & 60.16 & \bf 93.68 & \bf 93.55 & \bf 93.61 & 71.03 & 68.00\\
Qwen2.5-72B-Instruct & 22.06 & 12.34 & 15.83 & \bf 72.62 & \bf 44.60 & \bf 30.92 & \bf 36.52 & \bf 67.19 & 93.17 & 93.83 & 93.51 & \bf 72.84 & \bf 73.10 \\
\midrule
\multicolumn{14}{c}{\bf Consumer Staples}\\
\midrule
GPT-4o & \bf 27.43 & 10.90 & 15.60 & 71.77 & 41.91 & 18.51 & 25.68 & 65.98 & 91.27 & \bf 92.38 & 91.82 & 70.58 & 69.47 \\
\rowcolor{Gray}
Qwen2-72B-Instruct & 26.63 & \bf 16.92 & \bf 20.70 & \bf 73.83 & 32.41& 22.08 & 26.27 & 60.14 & 91.76 & 91.07 & 91.41 & 71.71 & 68.63 \\
Qwen2.5-72B-Instruct & 24.34 & 13.05 & 17.23 & 73.75 & \bf 44.79 & \bf 29.13 & \bf 35.30 & \bf 64.36 & \bf 92.20 & 92.29 & \bf 92.24 & \bf 73.00 & \bf 70.41 \\
\midrule
\multicolumn{14}{c}{\bf Energy}\\
\midrule
GPT-4o & 48.60 & 29.09 & 36.40 & 80.13 & \bf 55.80 & 38.54 & 45.59 & \bf 74.14 & \bf 95.40 & \bf 95.03 & \bf 95.21 & 79.14 &  78.12\\
\rowcolor{Gray}
Qwen2-72B-Instruct & 50.53 & \bf 47.49 & \bf 48.96 & 81.26 & 26.93 & 25.19 & 26.03 & 58.57 & 95.38 & 94.65 & 95.01 & 82.81 & 75.17\\
Qwen2.5-72B-Instruct & \bf 53.81 & 40.13 & 45.97 & \bf 81.39 & 52.88 & \bf 45.41 & \bf 48.87 & 73.90 & \bf 95.40 & \bf 95.03 & \bf 95.21 & \bf 85.10 & \bf 80.30\\
\midrule
\multicolumn{14}{c}{\bf Financials}\\
\midrule
GPT-4o & 24.13 & 8.68 & 12.77 & 73.81 & 41.34 & 29.62 & 34.52 & \bf 68.50 & \bf 95.49 & \bf 95.49 & \bf 95.49 & 68.59 & 70.37\\
\rowcolor{Gray}
Qwen2-72B-Instruct & \bf 29.62 & \bf 16.87 & \bf 21.50 & 71.37 & 28.38 & 24.31 & 26.19 & 61.41 & 94.47 & 90.82 & 92.61 & 72.61 & 68.52\\
Qwen2.5-72B-Instruct & 26.92 & 12.15 & 16.75 & \bf 75.02 & \bf 46.91 & \bf 36.71 & \bf 41.19 & 68.43 & 95.41 & 90.49 & 92.89 & \bf 72.70 & \bf 72.11\\
\midrule
\multicolumn{14}{c}{\bf Healthcare}\\
\midrule
GPT-4o & 24.00 & 9.25 & 13.36 & 73.28 & 41.51 & 25.72 & 31.76 & 67.57 & \bf 93.52 & \bf 93.52 & \bf 93.52 & 74.07 & 71.63\\
\rowcolor{Gray}
Qwen2-72B-Instruct & 28.67 & \bf 16.97 & \bf 21.32 & \bf 73.72 & 22.22 & 20.71 & 21.44 & 58.41 & 93.41 & 89.48 & 91.40 & 75.59 & 69.57\\
Qwen2.5-72B-Instruct & \bf 29.07 & 12.76 & 17.73 & 73.37 & \bf 44.52 & \bf 37.54 & \bf 40.73 & \bf 68.43 & 93.37 & \bf 93.52 & 93.45 & \bf 76.43 & \bf 72.74\\
\midrule
\multicolumn{14}{c}{\bf Industrials}\\
\midrule
GPT-4o & 27.15 & 10.56 & 15.20 & 72.13 & 43.81 & 23.08 & 30.23 & 63.86 & 92.22 & 89.85 & 91.02 & 68.74 & 68.28\\
\rowcolor{Gray}
Qwen2-72B-Instruct & 28.93 & \bf 18.15 & \bf 22.30 & 73.33 & 31.95 & 23.60 & 27.15 & 60.25 & 92.52 & 92.37 & 92.44 & 71.75 & 68.50\\
Qwen2.5-72B-Instruct & \bf 29.35 & 13.53 & 18.52 & \bf 73.94 & \bf 50.84 & \bf 36.27 & \bf 42.34 & \bf 65.46 & \bf 92.53 & \bf 92.53 & \bf 92.53 & \bf 73.67 & \bf 71.05\\
\midrule
\multicolumn{14}{c}{\bf Information Technology (IT)}\\
\midrule
GPT-4o & 23.82 & 11.02 &15.07 & 75.28 & 46.21 & 23.66 & 31.30 & 70.25 & 92.25 & \bf 93.07 & 92.66 & 73.47 & 73.05\\
\rowcolor{Gray}
Qwen2-72B-Instruct & \bf 28.78 & \bf 21.13 & \bf 24.37 & \bf 75.58 & 29.32 & 20.02 & 23.79 & 62.55 & 92.59 & 91.72 & 92.15 & 75.30 & 71.25\\
Qwen2.5-72B-Instruct & 27.38 & 16.29 & 20.42 & 75.44 & \bf 49.45 & \bf 32.00 & \bf 38.85 & \bf 72.26 & \bf 92.66 &92.77 & \bf 92.72 & \bf 76.67 & \bf 74.80\\
\midrule
\multicolumn{14}{c}{\bf Materials}\\
\midrule
GPT-4o & 30.17 & 13.71 & 18.85 & 70.63 & 42.80 & 27.95 & 33.81 & 64.58 & 91.34 & 91.56 & 91.45 & 73.04 & 69.5\\
\rowcolor{Gray}
Qwen2-72B-Instruct & \bf 31.63 & \bf 25.00 & \bf 27.92 & \bf 71.47 & 38.03 &29.87 & 33.46 & 61.42 & \bf  91.58 & \bf 91.80 & \bf 91.69 & 75.19 & 69.51\\
Qwen2.5-72B-Instruct & 29.36 & 18.54 &22.73 & 70.53 & \bf 49.67 & \bf 36.86 & \bf 42.32 & \bf 65.96 & 91.56 & 91.56 & 91.56 & \bf 76.42 & \bf 70.94\\
\midrule
\multicolumn{14}{c}{\bf Real Estate}\\
\midrule
GPT-4o & \bf 66.66 & 38.16 & 48.54 & 82.23 & 38.80 & 40 & 39.39 & \bf 74.50 & 89.70 & \bf 93.84 & 91.72 & 78.88 & \bf 79.17\\
\rowcolor{Gray}
Qwen2-72B-Instruct & 48.00 & 36.64 & 41.55 & 82.85 & 16.92 &16.92 & 16.92 & 66.50 & \bf 93.84 & \bf 93.84 & \bf 93.84 & 82.75 & 77.35\\
Qwen2.5-72B-Instruct & 61.29 & \bf 43.51 & \bf 50.89 & \bf 83.54 & \bf 48.27 & \bf 43.07 & \bf 45.52 & 67.25 & \bf 93.84 & \bf 93.84 & \bf 93.84 & \bf 83.75 & 79.10\\
\midrule
\multicolumn{14}{c}{\bf Utilities}\\
\midrule
GPT-4o & \bf 57.77 & 43.33 & 49.52 & 79.25 & 39.28 & 22.44 & 28.57 & 63.33 & \bf 83.67 & \bf 83.67 & \bf 83.67 & 82.00 & 76.42\\
\rowcolor{Gray}
Qwen2-72B-Instruct & 48.64 & \bf 60.00 & \bf 53.73 & \bf 80.25 & 22.72 & 20.40 & 21.50 & 62.33 & \bf 83.67 & \bf 83.67 & \bf 83.67 & 85.67 & 77.35\\
Qwen2.5-72B-Instruct & 53.70 & 48.33 & 50.87 & 79.50 & \bf 65.95 & \bf 63.26 & \bf 64.58 & \bf 69.33 & \bf 83.67 & \bf 83.67 & \bf 83.67 & \bf 86.33 & \bf 78.78\\
\bottomrule
\end{tabular}
}
\caption{Performance of LLMs on three evaluation tasks with different GICS sectors. To save space, we only report the performance of three LLMs.}
\label{tab:result_sector}
\end{table*}

\section{Annotation Guidelines}
\label{apx:annotation_guideline}

\begin{figure*}[!th]
\centering
\includegraphics[width=1.0\linewidth]{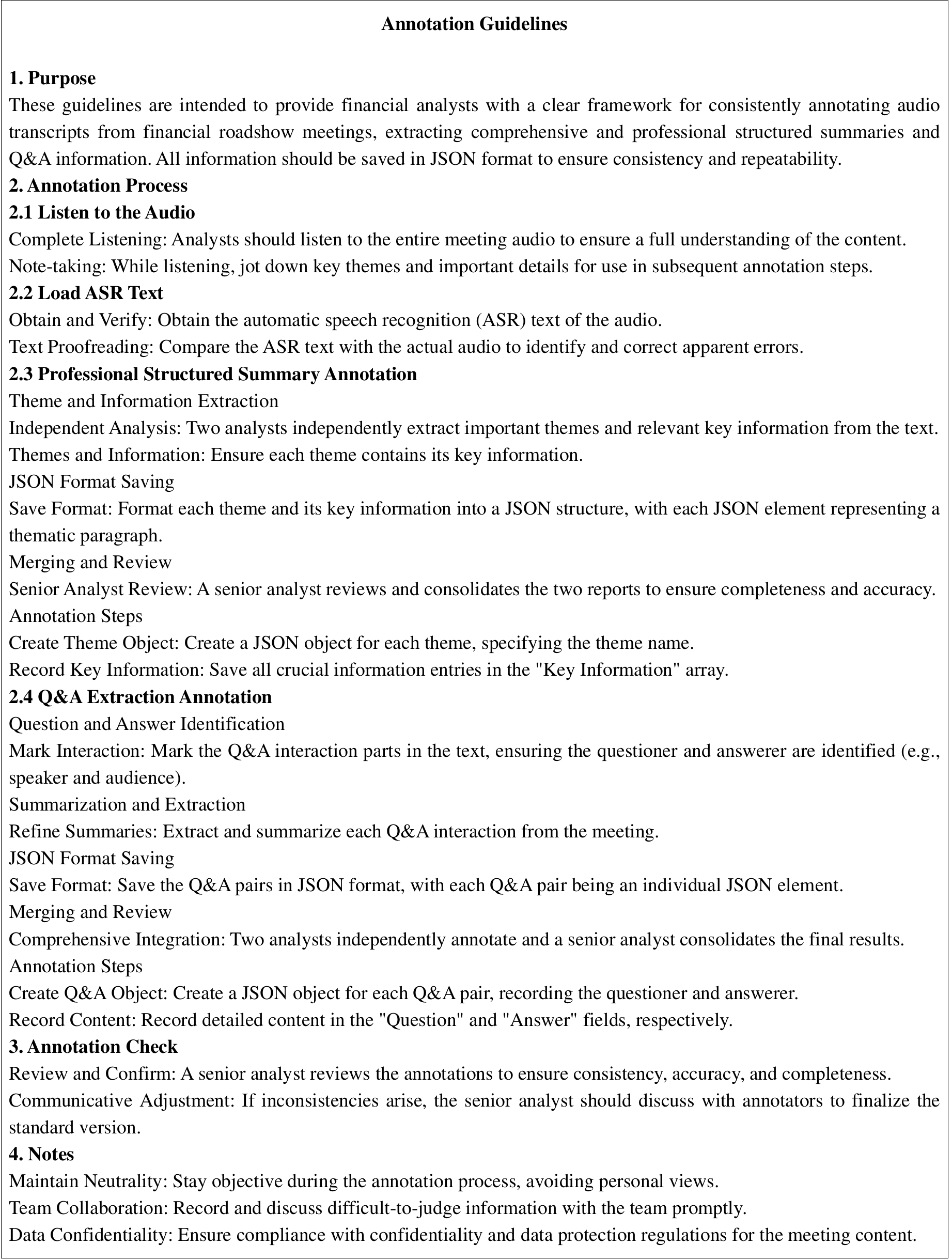}
\caption{Annotation guidelines of \texttt{M$^3$FinMeeting}. }
\label{fig:annotation_guideline}
\end{figure*}

Figure~\ref{fig:annotation_guideline} shows the annotation guideline of \texttt{M$^3$FinMeeting}.

\end{document}